\documentclass[final]{clv2025}

\jvol{vv}
\jnum{nn}
\jyear{2025}

\usepackage{hyperref}       
\usepackage{url}            
\usepackage{amsmath}
\usepackage{booktabs}
\usepackage{graphicx}
\usepackage{subfigure}        
\usepackage{tabularx}       
\usepackage{makecell}
\usepackage{multirow}
\usepackage{multicol}
\usepackage{xcolor}         
\usepackage{tabu}
\usepackage{tabularray}
\usepackage{bbding}         
\usepackage{arydshln}
\usepackage{CJKutf8}

\usepackage{amssymb}
\usepackage{pifont}

\runningtitle{TokAlign++}
\runningauthor{Li et al.}

\begin{document}

\title{TokAlign++: Advancing Vocabulary Adaptation via Better Token Alignment}

\author{Chong Li, Yingzhuo Deng, Wen Yang, 
        Jiajun Zhang\thanks{Corresponding authors}, Chengqing Zong}

\affilblock{
    \affil{State Key
 Laboratory of Multimodal Artificial Intelligence Systems, Institute of Automation, CAS, Beijing, China}
    \affil{School of Artificial Intelligence, University of Chinese Academy of Sciences, Beijing, China\\\quad \email{jjzhang@nlpr.ia.ac.cn}}
}

\maketitle

\begin{abstract}
Tokenization is a foundational step in the text process of Large Language Models (LLMs). 
Texts must be first tokenized into token IDs, which are then input to LLMs. 
Inefficient tokenization results in long token-ID sequences and will slow down the training and inference of LLMs. 
The fine-grained knowledge transfer between LLMs, like token-level distillation, is also impeded by the mismatch in vocabulary. 
To bridge this gap, we introduce a method named TokAlign++ to improve vocabulary adaptation performance by learning better token alignment lexicon. 
The source and target vocabularies are taken as two different languages, and the bilingual token alignment lexicon is learned from monolingual token representations. 
Model parameters are rearranged following this bilingual lexicon for new vocabulary, and progressively fine-tuned for adaptation. 
Experimental results on 15 languages show that our method boosts the multilingual text compression rates and preserves most of the multilingual ability of vanilla models. 
It costs as few as 1k steps to restore the performance of the vanilla model. 
After unifying vocabularies between vanilla models, token-level distillation remarkably improves the base model with only 235M tokens.
\end{abstract}

\section{Introduction}

\begin{figure*}[ht]
    \centering
    \subfigure[]{\includegraphics [scale=0.78]{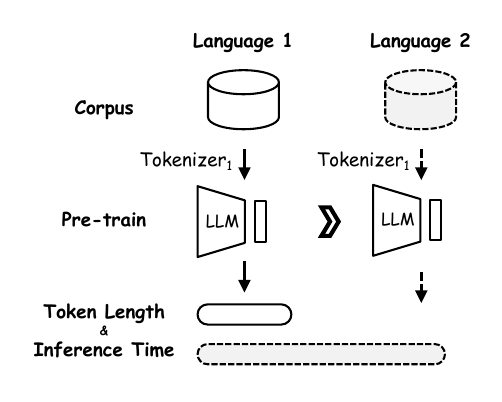}\label{fig:process}}
    \subfigure[]{\includegraphics [scale=0.46]{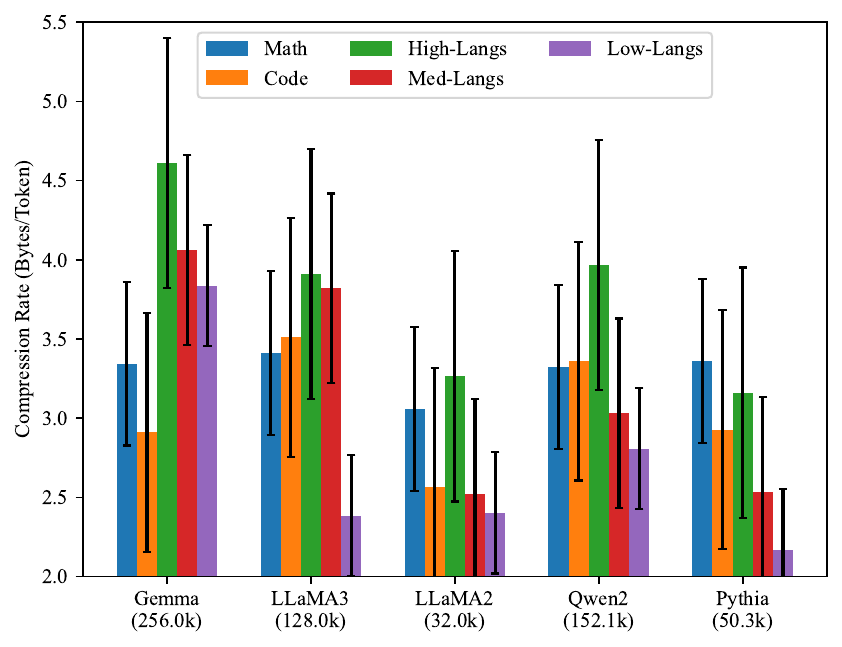}\label{fig:tok_compress}}
    \caption{\label{fig:demo}Motivation of TokAlign++. (a) When large language models learn new languages, the length of tokens and inference time cost often increase for the lower compression rate of vanilla Tokenzier on the new language. (b) Many large language models have their own tokenizer with different sizes. The compression rates of code and low-resource language texts often lag behind other domains.}
\end{figure*}

Tokenization is first proposed in the neural machine translation task to address the out-of-vocabulary problem and compress text~\cite{sennrich-etal-2016-neural, wu2016wordpieces, kudo-2018-subword}. 
It now serves as a foundational step for large language models~\cite{radford2018gpt, radford2019gpt2, brown-et-al-2020-gpt3, touvron2023llama, openai2023gpt4, glm2024chatglm, yang2024qwen2, yang2025qwen3} to tokenize text input into tokens. 
As shown in Figure \ref{fig:process}, the compression rate of vanilla tokenizers is often low on new languages or domain corpora, which increases the time and memory cost during inference and training of LLMs. 
Take LLaMA3 as an example, given the same byte size of text, Armenian text is 3.95$\times$ longer than English text after tokenization~\cite{li-etal-2025-tokalign}. 
Thus, it is costly to adopt LLaMA3 to deal with some low-resource languages, e.g., Armenia, texts. 
In addition, the specific pre-training corpus and method of each LLM contribute to its distinct strengths and weaknesses. 
Figure \ref{fig:tok_compress} illustrates five large language model vocabulary sizes and their compression rates on five domains. 
We can find that the vocabulary size varies across large language models. 
The deep knowledge transfer between them, e.g., token-level distillation and ensemble~\cite{xu-etal-2024-bridging, lu2024survey}, is also hindered by the mismatch of vocabulary. 
Therefore, it is important to investigate efficient vocabulary adaptation methods for the huge cost of re-training LLM with a new tokenizer. 

To address the above problems, heuristic methods~\cite{gee-etal-2022-fast, minixhofer-etal-2022-wechsel, downey-etal-2023-embedding, dobler-de-melo-2023-focus, liu-etal-2024-ofa} are proposed to better initialize the parameters of the target vocabulary for new tokenizer. 
Given each target token, FVT first tokenizes it using the source tokenizer and adopts the average parameter of the generated source tokens to initialize its parameters~\cite{gee-etal-2022-fast}. 
\citet{downey-etal-2023-embedding} classifies each source token by script, and then calculates the mean and standard deviation of embedding parameters using all tokens in each script. 
The parameters of some new tokens are sampled from the normal distribution with the mean and standard deviation of their own script, while the other tokens that do not belong to any scripts are randomly initialized. 
Focus~\cite{dobler-de-melo-2023-focus} relies on common tokens in the source and target vocabularies to initialize new parameters. 
It first trains a fastText~\cite{bojanowski2017fasttext} vector for each target token, and estimates the similarities between each target token and all common tokens. 
Then, the embedding of each target token is initialized by a weighted sum of common token embeddings, where the similarity after sparsemax~\cite{martins2016sparsemax} is taken as the weight. 
However, the performance of heuristic methods is not reliable, depending on many factors, like the number of overlapping tokens for Focus. 

On the other hand, some prior research studies~\cite{marchisio-etal-2023-mini, chen2023improving, minixhofer-2024-zett} attempt to modify the training process of the language model for efficient vocabulary adaptation. 
\citet{chen2023improving} found that actively forgetting, i.e., resetting, the embedding parameters every K steps can boost the convergence of new vocabulary adaptation. 
\citet{minixhofer-2024-zett} introduced a process to train a hyper-network for predicting the parameters of new vocabulary. 
It will cost a significant amount of computation for more parameters added.

\begin{figure*}[t]
\centering
\includegraphics [width=0.6\textwidth]{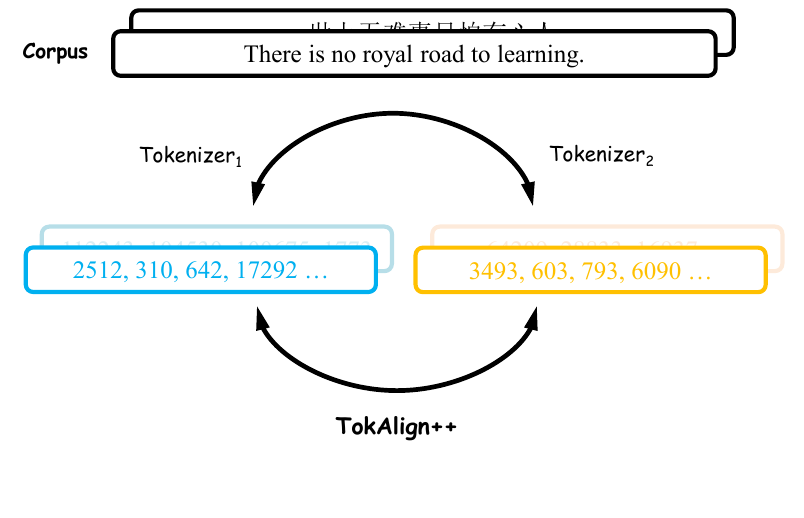}
\caption{\label{fig:motivation} TokAlign++ aims to bridge these corpora through aligning the similar token ID pairs.}
\end{figure*}

Large language models generally follow a common pre-training process: their tokenizer first tokenizes the pre-training corpus, then inputs it into the model for training. 
As shown in Figure \ref{fig:motivation}, the semantic and syntactic information is still reserved in the token-token co-occurrence, although the different tokenizers bring different token ID sequences for the same corpus. 
Motivated by this process, we propose a method named \textbf{TokAlign++} to efficiently replace the vocabulary of large language models. 
TokAlign++ aligns the source tokens and the tokens in the target vocabulary based on the global token-token co-occurrence matrix~\cite{pennington-etal-2014-glove}, and learns a token-token alignment matrix using bilingual lexicon induction methods~\cite{artetxe-etal-2018-robust}. 
Moreover, TokAlign++ introduces an unsupervised method to obtain token representation for token alignment from the vanilla LLM without training representation from scratch. 
There are two metrics designed to evaluate the token-token alignment matrix from the view of n-gram token matching and semantic similarity. 
Given the learned alignment matrix, the parameters of the most similar source token for each new target token are used for the initialization of the embedding and language modeling head of LLM, which is called ``\textit{LM\_Head}'' in the Transformers~\cite{wolf2019huggingface}. 
After new parameter initialization, we divide the vocabulary adaptation process into a progressive two-stage fine-tuning procedure to boost and stabilize the convergence. 

Given a multilingual vocabulary to replace, LLM preserves most of its performance after initialization using TokAlign++, decreasing the perplexity (PPL) on a 15-language corpus to 76.3 and outperforming strong baseline method ZeTT~\cite{minixhofer-2024-zett} (1.7$e^2$ PPL), and improves 47.6\% compression rates across 15 languages on average. 
It is noted that our method does not require an additional hyper-network to initialize parameters, which costs hundreds of GPU hours to train~\cite{minixhofer-2024-zett}. 
The two-stage vocabulary adaptation process of TokAlign++ brings 4.08$\times$ faster convergence than strong baseline methods. 
Experimental results on models with multiple parameter amounts demonstrate that our method costs as few as 1k steps to recover the performance of vanilla models on the general domain. 
In addition, standardizing the vocabulary across LLMs enhances deep knowledge transfer like token-level distillation, resulting in a 4.4\% improvement compared to sentence-level distillation using the same corpus.
Token-level distillation from a capable LLM enables the 1B model to achieve in-context abilities on par with a vanilla 7B model. 
Our contribution can be summarized as follows:

\begin{itemize}
    \item We propose an unsupervised method, TokAlign++, to align token IDs between two vocabularies, and replace the vocabulary of LLMs from the token-token co-occurrence view. It not only brings a good initialization for new vocabulary, but also costs as few as 1k steps to recover model performance. 
    \item We introduce two metrics to evaluate the performance of the token-level alignment matrix learned from the surface string matching and semantic similarity perspective. 
    Quantitative analysis demonstrates that they are linearly proportional to the initial loss of language model pre-training process. 
    \item Experimental results on ten datasets across fifteen languages show that our method promotes the cross-lingual knowledge transfer among multiple languages and deep knowledge transfer between models, like token-level distillation. 
\end{itemize}

As an extension of our conference paper in~\cite{li-etal-2025-tokalign}, this paper conducts the following improvements:
1) Propose an effective method to obtain the token representation without the need to train a token representation from the tokenized corpus. 
It achieves a result in vocabulary adaptation experiments comparable to the one of training token representations from scratch. 
2) Introduce bilingual lexicon induction methods to significantly improve the token alignment results, achieving new state-of-the-art performance on the vocabulary adaptation task with only 20\% training steps of our conference method. 
3) Design a bi-directional evaluation method for the token alignment stage, which is also reliable to evaluate the token-level alignment matrix. 
4) Conduct extensive experiments to show the effectiveness of our method, and a thorough ablation study to investigate the contribution of each module. 

The rest of this paper is organized as follows: Section \ref{sec:related_work} reviews the previous research work in related domains. 
Section \ref{sec:task} formulates the investigated research question. 
Section \ref{sec:method} introduces each module and the general process of TokAlign++. 
Section \ref{sec:exp} reports the experimental results to demonstrate the effectiveness and application of our method. 
Section \ref{sec:analysis} further provides valuable findings and an ablation study. 
Section \ref{sec:limitations} summaries the limitations of TokAlign++. 
Section \ref{sec:conclusion} draws a conclusion for this paper.

\section{Related Work}
\label{sec:related_work}
We will briefly review the three lines of related work: word representation, vocabulary adaptation, and bilingual lexicon induction. 

\subsection{Word Representation}

The training of word representation is based on the distributional semantic hypothesis~\cite{bengio2003neural}. 
There are many studies on improving the training efficiency~\cite{mikolov2013cbow,mikolov2013skip,pennington-etal-2014-glove} of word representation, which is used to initialize the embedding parameter of deep neural networks like LSTM~\cite{hochreiter1997lstm}, GRU~\cite{chung2014empirical}, and CNN~\cite{kim-2014-convolutional}. 
In order to learn a better word representation, some works attempt to incorporate more information, such as character-level morphological information~\cite{soricut-och-2015-unsupervised, cui2015knet,  bojanowski2017fasttext} or multi-modal information~\cite{wang2018learning, wang-etal-2018-associative}. 
On the other hand, dynamic word representations based on pre-trained language models like ELMo~\cite{peters-etal-2018-elmo}, BERT~\cite{devlin-etal-2019-bert}, and GPT~\cite{radford2018gpt} demonstrate better performance on the downstream NLP tasks. 
Our study trains the distributed representation for each token to align the token IDs between the source and target vocabularies. 

\subsection{Vocabulary Adaptation}
The training and inference of LLMs often slow down on the unseen domain or language corpus due to the inefficiency of the original tokenizer. 
Many vocabulary adaptation methods~\cite{tran2020english, wang-etal-2020-extending, chau-etal-2020-parsing, de-vries-nissim-2021-good, pfeiffer-etal-2021-unks, gee-etal-2022-fast, downey-etal-2023-embedding, zeng-etal-2023-greenplm, yamaguchi-etal-2024-empirical, minixhofer-etal-2022-wechsel, liu-etal-2024-ofa, dobler-de-melo-2023-focus, minixhofer-2024-zett, li-etal-2025-tokalign} have been proposed to address this problem. 
They first initialize new embedding-related parameters for the target vocabulary $\mathcal{V}_t$, then continue to train these new parameters, where the former is more important than the latter. 
These methods can be divided into two types according to whether they use heuristics.
Heuristic-based methods often incorporate surface~\cite{gee-etal-2022-fast} or semantic information~\cite{dobler-de-melo-2023-focus} into the parameter initialization process. 
\citet{gee-etal-2022-fast} initialized each target token $t \in {\mathcal{V}_t}$ with the mean of embeddings for the token sequence split by the source tokenizer. 
However, \citet{mundra-etal-2024-empirical} found that even simple heuristic-based methods, e.g., taking the mean of old embedding parameters to initialize the parameters of new tokens in the target vocabulary, perform comparably with some sophisticated methods like OFA~\cite{liu-etal-2024-ofa}. 
On the other hand, heuristic-free methods train a hypernetwork to predict the new embedding for the target vocabulary~\cite{pinter-etal-2017-mimicking, schick-schutze-2019-attentive, schick-schutze-2020-bertram, minixhofer-2024-zett} or improve the training process to boost the vocabulary adaptation~\cite{marchisio-etal-2023-mini, chen2023improving}. 

The most similar vocabulary adaptation methods to our work are the heuristic-based method WECHSEL~\cite{minixhofer-etal-2022-wechsel} and our conference method TokAlign~\cite{li-etal-2025-tokalign}. 
WECHSEL estimates the similarity between tokens through tokenizing all words in the provided bilingual dictionary and the weighted sum of additional word representations. 
On the contrary, our method can align source and target tokens in an unsupervised way without a bilingual dictionary and additional trained representations. 
As for our conference method, TokAlign, we further propose an unsupervised method to obtain token representation and improve its token alignment module, which brings significantly better zero-shot vocabulary performance and faster convergence. 

\subsection{Bilingual Lexicon Induction}
Given two monolingual corpora in the source and target languages or a machine translation parallel corpus, \textbf{B}ilingual \textbf{L}exicon \textbf{I}nduction(\textbf{BLI}) methods intend to extract word translation pairs between two languages. 
There are three types of BLI methods: statistical-based methods, linear mapping-based methods, and non-linear mapping-based methods. 
Statistical-based BLI methods~\cite{haghighi-etal-2008-learning, vulic-etal-2011-identifying, dyer-etal-2013-simple, wang-zong-2013-large, huang2016wordalign, e-zhou-2022-math} rely on the statistical analysis of the corpus to conduct the word-level alignment. 
For example, \citet{haghighi-etal-2008-learning} introduced a generative model based on canonical correlation analysis to extract translation pairs. 
After mathematical analysis of recurring patterns in the corpus, \citet{e-zhou-2022-math} adopted a Markov semantic model to build the semantic representations of words, then conducted automated word translation with these representations. 
As for the linear mapping-based BLI methods~\cite{vulic-moens-2015-bilingual, vulic-korhonen-2016-role, upadhyay-etal-2016-cross, lample2018word, artetxe-etal-2018-robust, joulin-etal-2018-loss, hoshen-wolf-2018-non, jawanpuria-etal-2019-learning, artetxe-etal-2019-bilingual, patra-etal-2019-bilingual, glavas-etal-2019-properly, taitelbaum-etal-2019-multilingual, ren-etal-2020-graph, zhao-etal-2020-semi, zhao-etal-2020-relaxed, wang-etal-2021-multi, peng-etal-2021-cross, li-etal-2022-improving, tian-etal-2022-rapo, ding-2024-enhancing, hu-xu-2024-dm}, word embeddings are linearly projected into a shared high-dimensional space where the nearest target word for each source word is taken as its translation in the target language. 
The difference between methods mostly lies in the training method of mapping matrices. 
For example, \citet{artetxe-etal-2018-robust} used self-learning to learn high-quality bilingual embedding mappings, while \citet{li-etal-2022-improving} exploited contrastive learning to train bilingual embedding mapping. 
Non-linear mapping-based BLI methods applied non-linear transformation approaches like instance-based mapping~\cite{glavas-vulic-2020-non} or non-linear autoencoder~\cite{mohiuddin-etal-2020-lnmap} to learn bilingual word embeddings. 
In this study, we take the token embeddings of source and target vocabulary as bilingual word embeddings, and use the bilingual lexicon induction method to improve the source-target token alignment performance. 

\section{Task Formulation}
\label{sec:task}
Given an LLM with the source vocabulary ${\mathcal{V}_s}$ and a target vocabulary ${\mathcal{V}_t}$ for the new tokenizer ${\mathcal{T}_t}$, the vocabulary adaptation method aims to replace the source vocabulary ${\mathcal{V}_s}$ of the LLM to the target vocabulary ${\mathcal{V}_t}$ while largely preserving its general performance. 
It often consists of two stages: parameter initialization for the new vocabulary and new parameter adaptation. 
The latter stage can be ignored if the performance of the model after initialization is good enough, which is called the zero-shot tokenizer transfer method~\cite{minixhofer-2024-zett}.

\begin{figure*}[ht]
\centering
\includegraphics [width=1\textwidth]{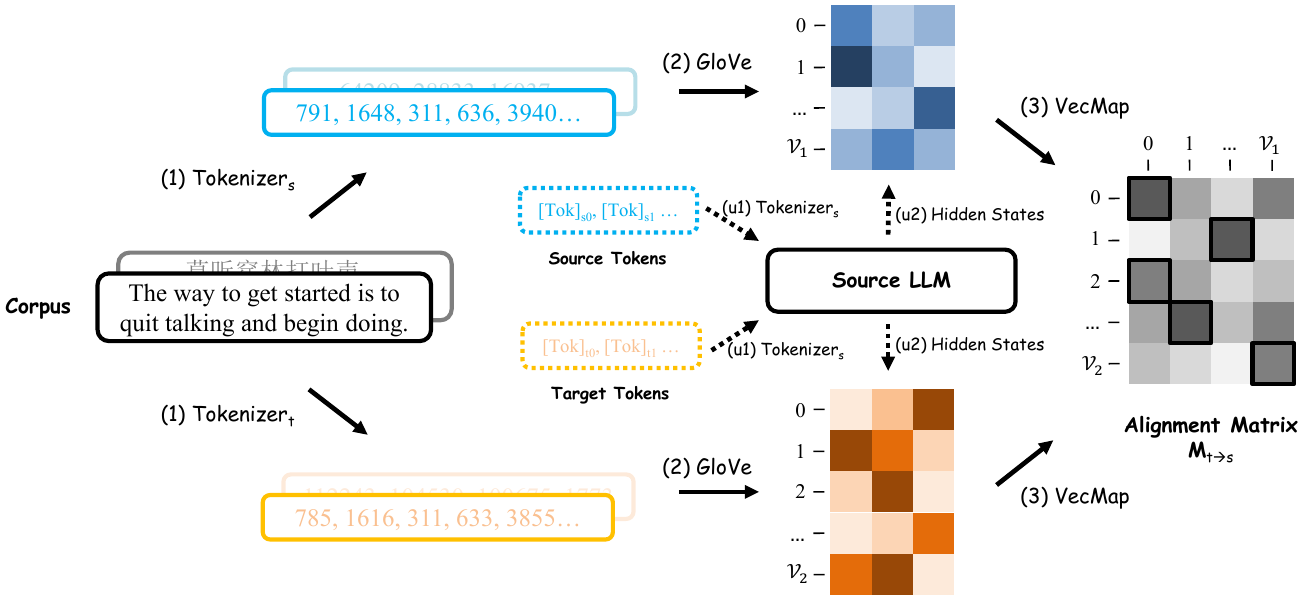}
\caption{\label{fig:method} Illustration of TokAlign++ to align token IDs from different vocabularies. We train token representations on the tokenized corpus or obtain the token representations from the source large language model (u1 and u2 steps), and align token IDs using the bilingual lexicon induction method VecMap. It is noted that the IDs of tokens belonging to both vocabularies are directly replaced without alignment.}
\end{figure*}

\section{Methodology}
\label{sec:method}
Figure \ref{fig:method} illustrates the first three steps of TokAlign++ to align tokens in source and target vocabularies. 
For token representation learning, we introduce two methods that training from the tokenized corpus or obtaining from the hidden states of a large language model for tokens. 
For token alignment, we adopt the bilingual lexicon induction methods to boost the initialization results. 
As shown in Figure \ref{fig:evalTune}, there are two metrics designed to evaluate the performance of the learned token alignment matrix and two-stage vocabulary adaptation process for new initialized parameters. 

\subsection{Vocabulary Alignment}
In order to obtain the token representation for alignment, we provide the following methods:

\paragraph{Train from Scratch} A comprehensive and diverse training corpus plays a vital role in developing well-trained token representations. 
It is hard to train the representation of tokens in the tail of vocabulary on an unbalanced corpus. 
Therefore, there are three source corpora: math corpus Proof-Pile-2 (30\%)~\cite{azerbayev2024llemma}, code corpus The Stack (30\%)~\cite{kocetkov2023stack}, and multilingual corpus CulturaX (40\%)~\cite{nguyen-etal-2024-culturax} used in this work. 
The mixed corpus is first tokenized by source and target tokenizers, resulting in multiple token ID sequences. 
We set the amount of tokens used for training token representations to 1B, which is further investigated in Section \ref{sec:ablation_token_rep}. 
GloVe~\cite{pennington-etal-2014-glove} is chosen to train the token representations on the tokenized corpus. 
The primary reason is that GloVe incorporates global statistical information from the entire corpus, unlike sliding window approaches such as CBOW~\cite{mikolov2013cbow, mikolov2013skip, bojanowski2017fasttext}, which focus on the local context. 
The settings to train GloVe vectors are introduced in Section \ref{sec:exp_settings}. 

\paragraph{Extract from LLM} Another method obtains token representations from the hidden states of the source large language model (u1 and u2 steps in Figure \ref{fig:method}).
It reduces the need to download and tokenize a training corpus, and saves the cost of disk storage and CPU computation. 
Specifically, source tokens and target tokens are input into the source large language model, and then the last hidden states of the source LLM are taken as the token representations for alignment. 

After obtaining token representations, we treat the token alignment task as the \textbf{B}ilingual \textbf{L}exicon \textbf{I}nduction(\textbf{BLI}) task, and adopt the BLI method VecMap~\cite{artetxe-etal-2018-robust} to align token IDs in source and target vocabularies. 
It should be noted that the IDs of tokens shared across both vocabularies are substituted directly without any alignment process. 
The learned token-token alignment matrix $M_{t\to s}$ saves the pair-wise similarity of each target token and source token, which can be used for each source/target token to find the most similar token from the target/source vocabulary. 
We choose the \textbf{C}ross-domain \textbf{S}imilarity \textbf{L}ocal \textbf{S}caling(\textbf{CSLS})~\cite{lample2018word} to quantify the similarity of two representations for avoiding the hubness problem that many points share the same few points as their nearest neighbors in the high-dimensional space~\cite{radovanovic2010hub, radovanovic2010on, dinu2015hub}. 
To be specific, it is computed for two mapped embeddings $x$ and $y$ as below:
\begin{equation}
    \text{CSLS}(x, y) = 2\text{cos}(x,y) - \text{r}_\text{T}(x) - \text{r}_\text{S}(y)
\end{equation}
where $\text{r}_\text{T}(x)$ and $\text{r}_\text{S}(y)$ are the average cosine similarity of $x$ and $y$ for their $k$, which is set to 10 following \citet{lample2018word}, nearest neighbors in the source and target vocabularies, correspondingly. 

\subsection{Alignment Evaluation}
\begin{figure*}[t]
\centering
\includegraphics [width=0.95\textwidth]{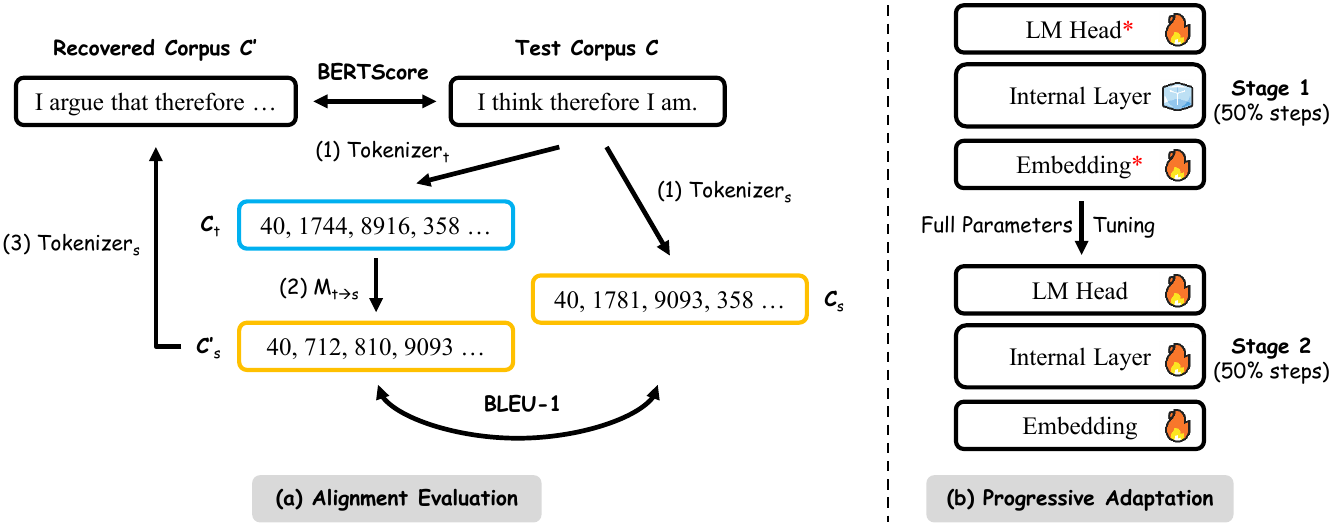}
\caption{\label{fig:evalTune} (a) We choose BLEU-1 and BERTScore to evaluate the performance of alignment matrix $M_{t\to s}$ (b) Embedding and LM$\_$Head are tuned at the first half part of the process, followed by full parameter tuning. {\color{red} \textbf{$\ast$}} indicates the parameter of each target token is first initialized from the most similar source token by alignment matrix $M_{t\to s}$.}
\end{figure*}

We propose two metrics to evaluate the token alignment matrix $M_{t\to s}$ from the view of token ID matching and semantic similarity, which is shown in Figure \hyperref[fig:evalTune]{4(a)}. 

The test corpus $\mathcal{C}$ is first tokenized by the source and target tokenizers, resulting in $\mathcal{C}_s$ and $\mathcal{C}_t$, respectively. 
Then, each token ID from the corpus $\mathcal{C}_t$ is changed to its most similar source token ID by the alignment matrix $M_{t\to s}$, switching to the corpus $\mathcal{C}_{s}^{'}$. 
In terms of token ID matching, a higher BLEU-1 score between $\mathcal{C}_{s}^{'}$ and the corpus $\mathcal{C}_s$ (tokenized by Tokenizer${}_{s}$) indicates a better alignment matrix $M_{t\to s}$. 

We also introduce a metric from the semantic evaluation perspective. 
It converts the source token ID corpus $\mathcal{C}_{s}^{'}$ back into readable text corpus $\mathcal{C}^{'}$ using Tokenizer${}_{s}$. 
The semantic similarity of $\mathcal{C}^{'}$ to the original corpus $\mathcal{C}$ is then measured via BERTScore. 
The better alignment matrix $M_{t\to s}$ learned retains more semantics in the test corpus $\mathcal{C}$, leading to a higher BERTScore between the recovered $\mathcal{C}^{'}$ and $\mathcal{C}$. 

On the other hand, it is noted that the token alignment can be evaluated from the source token to the target token direction. 
Specifically, the token alignment matrix $M_{s\to t}$ is applied to the tokenized corpus $\mathcal{C}_{s}$, resulting in the converted target token corpus $\mathcal{C}_{t}^{'}$. 
Then BLEU-1($\mathcal{C}_{t}$, $\mathcal{C}_{t}^{'}$) is used to quantify the performance of token alignment. 
The BERTScore between $\mathcal{C}$ and $\mathcal{C}^{'}$, which is recovered from the $\mathcal{C}_{t}^{'}$ by Tokenizer${}_{t}$, can be applied to evaluate in the same way. 

\subsection{Progressive Adaptation}
For the parameter initialization, the new parameters of each target vocabulary token are set to the ones of its most similar source token according to the alignment matrix $M_{t\to s}$. 
Our results in Section \ref{sec:res} show that these re-arranged embeddings and LM$\_$Head serve as an effective initialization for the new model. 
We design a two-stage tuning process for an LLM to adapt to the new vocabulary, which is shown in Figure \hyperref[fig:evalTune]{4(b)}.
To reduce loss spike and enhance the training stability, the vocabulary-related parameters (re-arranged embedding and LM$\_$Head) are fine-tuned first. 
The remaining parameters in the internal layers are jointly fine-tuned during the last half of the process.

\section{Experiments}
\label{sec:exp}
\subsection{Experiments Settings}
\label{sec:exp_settings}

\paragraph{Large Language Models} Our research is mainly built upon the Pythia~\cite{biderman2023pythia} models, which are fully open-source large language models. It should be clarified that our aim is not to set state-of-the-art large language model performance but to develop a more efficient vocabulary adaptation method for English-biased tokenizers like Pythia. 
LLaMA3~\cite{meta2024llama3} is taken as an example of models with an unknown pre-training corpus to replace vocabulary. 
We select the vocabularies of LLaMA2~\cite{touvron2023llama2}, LLaMA3~\cite{meta2024llama3}, Qwen2~\cite{yang2024qwen2}, and Gemma~\cite{team2024gemma} as the target for replacement, intending to transfer token-level knowledge from these capable large language models.

\paragraph{Corpus} The vanilla pre-training corpus Pile~\cite{gao2020pile} of Pythia is used in the two-step tuning stage to reduce the risk of distribution shift from the original training data. 
To assess the training corpus robustness in vocabulary alignment, we also adopt the Pile corpus in the vocabulary adaptation process of LLaMA3. 
For the evaluation of cross-lingual and cross-model transfer, the multilingual corpus CulturaX~\cite{nguyen-etal-2024-culturax} and training datasets of downstream tasks are also utilized in this study. 

\paragraph{Evaluation Datasets and Metrics} There are ten evaluation datasets, including reading comprehension ~\cite{clark-etal-2019-boolq} and commonsense reasoning tasks~\cite{clark2018arc, mihaylov-etal-2018-suit, zellers-etal-2019-hellaswag, ponti-etal-2020-xcopa, bisk-etal-2020-piqa, Sakaguchi-et-al-2020-winogrande}, selected in this work. 
In order to prevent the variability from the prompt and evaluation choices, we use the default prompt provided by the widely adopted language model evaluation harness framework~\cite{eval-harness}. 
For the evaluation metrics of large language models, perplexity on the validation corpus and in-context learning accuracy on the downstream tasks are used in this study. 

\paragraph{Implementation Details} The optimizer adopted in this work is AdamW~\cite{loshchilov2019adamw}, where $\beta_1 = 0.9$ and $\beta_2 = 0.999$. 
The learning rate for baseline methods is set to 5e-5 to reduce the loss spike. 
We adopt bf16 mixed precision training, ZeRO-1, and flash-attention to save GPU memory cost and speed up the training process~\cite{micikevicius2018mixed, rasley2020deepspeed, dao2022flash, dao2024flashattention2}. 
The reported results are averaged across three trials to account for experimental variance. 
Following \citet{biderman2023pythia}, the batch size is set to 2M tokens and the max sequence length is 2048. 
We reduce the number of tuning steps needed to 1k from the 5k steps in our conference paper due to the better initialized model parameters. 

\subsection{Baselines}
Existing vocabulary adaptation methods are compared as baseline methods:
\begin{itemize}
    \item \textbf{Random Initialization} reuses the old parameters if token $t \in \{\mathcal{V}_t\cap \mathcal{V}_s\}$, which belongs to both vocabularies. The token $t \in \{\mathcal{V}_t\setminus (\mathcal{V}_t \cap \mathcal{V}_s)\}$ is initialized by the default random initialization method of Huggingface Transformers. 
    \item \textbf{Random Permutation} also reuses the old parameters of shared tokens. For each token $t \in \{\mathcal{V}_t\setminus (\mathcal{V}_t \cap \mathcal{V}_s)\}$, it adopts the parameter of a randomly chosen token from the source vocabulary. 
    \item \textbf{Multivariate} samples from the multivariate Gaussian distribution with the mean and covariance of source embedding $\textit{E}_s$ to initialize each token $t \in \{\mathcal{V}_t\setminus (\mathcal{V}_t \cap \mathcal{V}_s)\}$. 
    \item \textbf{Mean} initializes all tokens $t \in \{\mathcal{V}_t\setminus (\mathcal{V}_t \cap \mathcal{V}_s)\}$ via the mean of source embedding $\textit{E}_s$. 
    \item \textbf{WECHSEL} \cite{minixhofer-etal-2022-wechsel} uses a linear transformation to convert source token embeddings into target token embeddings, achieved by tokenizing and recombining supplementary embeddings $\textit{W}^{s}$ and $\textit{W}^{t}$ aligned via a bilingual dictionary.
    \item \textbf{OFA} \cite{liu-etal-2024-ofa} factorizes the source model embeddings  $\textit{E}_s$ into a source coordinate $\textit{F}_s$ and a primitive embedding $\textit{P}$. Then, it applies the multilingual word embedding $\textit{W}$ to transform the source coordinate $\textit{F}_s$ into the target coordinate $\textit{F}_t$. Multiplying the assembled primitive embedding $\textit{P}$ by and target coordinate $\textit{F}_t$ results in the target embedding $\textit{E}_t$. 
    \item \textbf{Focus} \cite{dobler-de-melo-2023-focus} initializes the embedding parameters of token $t \in \{\mathcal{V}_t\setminus (\mathcal{V}_t \cap \mathcal{V}_s)\}$ using the weighted sum of the ones from the shared token $t \in \{\mathcal{V}_t\cap \mathcal{V}_s\}$. It largely depends on the amount of shared tokens, and performs poorly under the low overlapping percentage of $\mathcal{V}_t$ and $\mathcal{V}_s$. 
    \item \textbf{ZeTT} \cite{minixhofer-2024-zett} trains an additional hypernetwork $\textit{H}_{\theta}$ to generate the parameters for each token $t \in \mathcal{V}_t$. It costs a lot to train the added hypernetwork. 
\end{itemize}

\begin{table*}[htp]

\renewcommand\arraystretch{0.8}

\centering
\scriptsize

\setlength{\tabcolsep}{0.17mm}
\caption{\label{tab:multilingual_ppl} The normalized perplexity on the valid corpus of CulturaX. The perplexity is normalized to the vocabulary of Pythia following \citet{wei2023skywork}. 
``\textbf{High}'', ``\textbf{Medium}'', and ``\textbf{Low}'' indicates the available amount of linguistic resources. 
}
 \begin{tabular}{lcccccccccccccccc}
 
 \toprule[1.2pt]
  
  \multicolumn{1}{c}{ } & \multicolumn{5}{c}{\textbf{High}} & \multicolumn{5}{c}{\textbf{Medium}} & \multicolumn{5}{c}{\textbf{Low}} & \multicolumn{1}{c}{ } \\

  \cmidrule(r){2-6} \cmidrule(r){7-11} \cmidrule(r){12-16} \noalign{\smallskip}

 \multicolumn{1}{c}{\textbf{Model}} & \textbf{ar} & \textbf{de} & \textbf{en} & \textbf{ja} & \textbf{zh} & \textbf{bn} & \textbf{ko} & \textbf{th} & \textbf{uk} & \textbf{vi} & \textbf{ml} & \textbf{mn} & \textbf{ta} & \textbf{te} & \multicolumn{1}{c}{\textbf{ur}} & \multicolumn{1}{c}{\textbf{Avg} $\downarrow$} \\

\midrule[0.8pt]

$\text{Qwen2}_{\text{1.5B}}$
&$4.6$&$11.3$&$10.4$&$5.6$&$4.5$&$2.3$&$3.1$&$2.4$&$5.4$&$3.2$&$2.3$&$6.0$&$2.7$&$3.2$&$3.8$&$4.7$\\

\midrule[0.8pt]

$\text{Pythia}_{\text{1B}}$
&$7.4$&$15.8$&$\textbf{13.1}$&$8.8$&$12.8$&$3.3$&$5.4$&$4.0$&$6.4$&$5.9$&$2.6$&$8.0$&$2.8$&$3.1$&$5.2$&$7.0$\\[0.25em]\hdashline\\[-0.5em]

\multicolumn{17}{c}{\textit{Perplexity after initialization without any tuning.}} \\[0.3em]\hdashline\\[-0.4em]



$\text{\ w/\ }{\text{Focus}}$
&$4.8e^3$&$2.3e^5$&$1.7e^6$&$1.6e^4$&$9.8e^2$&$6.3e^4$&$7.6e^2$&$4.8e^3$&$1.7e^6$&$8.3e^2$&$2.1e^3$&$1.7e^5$&$4.2e^4$&$1.7e^5$&$2.0e^5$&$2.9e^5$\\ 

$\text{\ w/\ }{\text{ZeTT}}$
&$2.3e^2$&$6.6e^2$&$1.1e^2$&$1.9e^2$&$2.3e^2$&$1.1e^2$&$2.1e^2$&$42.1$&$3.2e^2$&$1.5e^2$&$1.1e^2$&$3.4e^2$&$31.2$&$2.0e^2$&$1.2e^2$&$2.0e^2$\\

$\text{\ w/\ }{\text{TokAlign}}$
&$1.5e^2$&$3.8e^2$&$1.1e^2$&$3.0e^2$&$1.3e^2$&$41.3$&$69.2$&$63.7$&$1.6e^2$&$50.4$&$20.2$&$3.8e^2$&$52.7$&$95.1$&$78.2$&$1.4e^2 $\\

$\text{\ w/\ }{\text{TokAlign++}}$
&$1.1e^2$&$3.0e^2$&$63.0$&$1.3e^2$&$95.8$&$20.2$&$49.0$&$25.5$&$1.3e^2$&$46.6$&$10.9$&$1.2e^2$&$12.2$&$18.1$&$37.7$&$78.4$\\[0.25em]\hdashline\\[-0.5em]

\multicolumn{17}{c}{\textit{Perplexity after 1k steps tuning on multilingual corpus.}}\\[0.3em]\hdashline\\[-0.4em]

$\text{\ w/\ }{\text{Focus}}$&$10.1$&$40.0$&$45.2$&$19.5$&$20.7$&$3.2$&$6.0$&$4.0$&$10.4$&$6.8$&$2.1$&$6.4$&$2.8$&$2.9$&$4.6$&$12.3$\\

$\text{\ w/\ }{\text{ZeTT}}$&$8.1$&$19.1$&$18.2$&$11.8$&$12.5$&$3.0$&$5.7$&$3.8$&$7.1$&$5.8$&$2.4$&$5.9$&$2.7$&$2.9$&$4.5$&$7.6$\\

$\text{\ w/\ }{\text{TokAlign}}$&$6.4$&$12.4$&$16.2$&$9.1$&$9.6$&$2.4$&$4.3$&$3.0$&$5.7$&$4.2$&$\textbf{2.0}$&$4.6$&$\textbf{2.2}$&$2.4$&$3.6$&$6.3$\\

$\text{\ w/\ }{\text{TokAlign++}}$&$\textbf{5.9}$&$\textbf{11.8}$&$14.6$&$\textbf{7.8}$&$\textbf{7.8}$&$\textbf{2.3}$&$\textbf{4.1}$&$\textbf{2.9}$&$\textbf{5.0}$&$\textbf{4.1}$&$\textbf{2.0}$&$\textbf{4.3}$&$\textbf{2.2}$&$\textbf{2.3}$&$\textbf{3.5}$&$\textbf{5.7}$\\

\midrule[0.8pt]

$\text{Qwen2}_{\text{7B}}$&$3.8$&$8.3$&$8.2$&$4.6$&$3.7$&$2.1$&$2.8$&$2.2$&$3.7$&$2.8$&$2.0$&$4.6$&$2.2$&$2.5$&$3.1$&$3.8$\\

\midrule[0.8pt]

 $\text{Pythia}_{\text{6.9B}}$
&$5.9$&$11.1$&$\textbf{10.5}$&$7.1$&$9.6$&$2.9$&$4.4$&$3.4$&$4.7$&$4.6$&$2.4$&$6.4$&$2.5$&$2.8$&$4.3$&$5.5$\\[0.25em]\hdashline\\[-0.5em]

\multicolumn{17}{c}{\textit{Perplexity after initialization without any tuning.}} \\[0.3em]\hdashline\\[-0.4em]
 

$\text{\ w/\ }{\text{Focus}}$
&$6.5e^3$&$2.1e^5$&$1.1e^6$&$1.9e^4$&$1.6e^3$&$2.2e^4$&$4.8e^2$&$3.0e^3$&$1.9e^6$&$7.7e^3$&$1.8e^3$&$1.3e^5$&$1.4e^4$&$1.2e^5$&$1.4e^5$&$2.5e^5$\\

$\text{\ w/\ }{\text{ZeTT}}$
&$1.3e^2$&$4.1e^2$&$67.6$&$1.4e^2$&$1.7e^2$&$56.1$&$98.3$&$35.7$&$1.9e^2$&$81.5$&$81.9$&$2.1e^2$&$27.5$&$83.7$&$94.5$&$1.3e^2$\\

$\text{\ w/\ }{\text{TokAlign}}$
&$1.5e^2$&$3.6e^2$&$98.8$&$2.6e^2$&$1.2e^2$&$38.6$&$52.8$&$43.1$&$1.6e^2$&$52.8$&$18.6$&$1.9e^2$&$41.9$&$89.9$&$53.5$&$1.1e^2$\\

$\text{\ w/\ }{\text{TokAlign++}}$
&$1.0e^2$&$2.9e^2$&$49.7$&$1.4e^2$&$85.0$&$21.7$&$39.2$&$25.0$&$1.5e^2$&$44.1$&$9.6$&$1.1e^2$&$12.1$&$14.1$&$29.7$&$74.2$\\[0.25em]\hdashline\\[-0.5em]

\multicolumn{17}{c}{\textit{Perplexity after 1k steps tuning on multilingual corpus.}} \\[0.3em]\hdashline\\[-0.4em]

$\text{\ w/\ }{\text{Focus}}$&$7.8$&$18.1$&$32.0$&$12.2$&$12.9$&$2.5$&$5.0$&$3.3$&$7.4$&$5.2$&$2.1$&$5.2$&$2.3$&$2.4$&$4.0$&$8.2$\\

$\text{\ w/\ }{\text{ZeTT}}$&$6.6$&$13.0$&$15.0$&$9.1$&$9.3$&$2.9$&$4.9$&$3.5$&$5.4$&$4.9$&$2.5$&$5.1$&$2.7$&$2.9$&$4.2$&$6.1$\\

$\text{\ w/\ }{\text{TokAlign}}$&$6.1$&$10.3$&$15.0$&$8.7$&$9.6$&$2.3$&$4.1$&$2.9$&$5.5$&$4.0$&$1.9$&$4.4$&$2.1$&$2.2$&$3.5$&$5.5$\\

$\text{\ w/\ }{\text{TokAlign++}}$&$\textbf{4.9}$&$\textbf{9.1}$&$11.6$&$\textbf{6.6}$&$\textbf{6.5}$&$\textbf{2.1}$&$\textbf{3.5}$&$\textbf{2.5}$&$\textbf{4.0}$&$\textbf{3.5}$&$\textbf{1.8}$&$\textbf{3.7}$&$\textbf{2.0}$&$\textbf{2.1}$&$\textbf{3.0}$&$\textbf{4.4}$\\

\midrule[0.8pt]

$\Delta\textbf{\ Length (\%) $\downarrow$}$&$-47.9$&$-27.9$&$-2.1$&$-46.8$&$-47.1$&$-58.5$&$-52.0$&$-61.1$&$-43.1$&$-53.3$&$-63.7$&$-29.0$&$-65.3$&$-59.6$&$-56.8$&$-47.6$\\

\bottomrule[1.2pt]
\end{tabular}
\end{table*}

\subsection{Overall Performance}
\label{sec:res}
\paragraph{Cross-lingual Transferring Experiments} When large language models are used in new languages or domains, more efficient tokenizers can accelerate the training and inference process tasks. 
Token co-occurrence enables aligning different language tokens with their semantically similar source vocabulary tokens, thereby boosting cross-lingual knowledge transfer. 
Therefore, we substitute the English-biased Pythia tokenizer with the Gemma tokenizer to assess the performance of our method in cross-lingual transfer scenarios. 

\begin{table*}[htp]

\renewcommand\arraystretch{0.8}

\centering
\scriptsize

\setlength{\tabcolsep}{0.38mm}
\caption{\label{tab:multilingual_0_shot} Zero-shot in-context learning results of cross-lingual transfer. 
}
 \begin{tabular}{lcccccccccccccccccccc}
 
 \toprule[1.2pt]
  
  \multicolumn{1}{c}{ } & \multicolumn{7}{c}{\textbf{XNLI}} & \multicolumn{5}{c}{\textbf{PAWS-X}} & \multicolumn{3}{c}{\textbf{XCOPA}} & \multicolumn{4}{c}{\textbf{XStoryCloze}} & \multicolumn{1}{c}{ } \\

  \cmidrule(r){2-8} \cmidrule(r){9-13} \cmidrule(r){14-16} \cmidrule(r){17-20} \noalign{\smallskip}
  
 \multicolumn{1}{c}{\textbf{Model}} & \textbf{en} & \textbf{de} & \textbf{zh} & \textbf{ar} & \textbf{th} & \textbf{vi} & \textbf{ur} & \textbf{de} & \textbf{en} & \textbf{ja} & \textbf{ko} & \textbf{zh} & \textbf{th} & \textbf{vi} & \textbf{ta} & \textbf{en} & \textbf{zh} & \textbf{ar} & \multicolumn{1}{c}{\textbf{te}} & \multicolumn{1}{c}{\textbf{Avg}} \\

\midrule[0.8pt]

$\text{Pythia}_{\text{1B}}$
&$\textbf{51.0}$&$37.8$&$42.6$&$35.9$&$34.8$&$37.0$&$34.7$&$49.6$&$49.3$&$54.8$&$54.9$&$52.9$&$54.0$&$53.2$&$\textbf{55.4}$&$64.3$&$48.6$&$\textbf{48.0}$&$52.9$&$48.0$\\[0.25em]\hdashline\\[-0.5em]

\multicolumn{21}{c}{\textit{Performance after initialization without any tuning.}} \\[0.3em]\hdashline\\[-0.4em]



$\text{\ w/\ }{\text{Focus}}$
&$33.8$&$34.6$&$34.7$&$35.4$&$33.8$&$34.4$&$33.9$&$48.6$&$50.4$&$45.7$&$44.8$&$47.3$&$52.4$&$48.6$&$54.4$&$44.7$&$47.8$&$46.4$&$49.4$&$43.2$\\ 

$\text{\ w/\ }{\text{ZeTT}}$
&$46.2$&$34.9$&$33.9$&$33.0$&$33.4$&$34.2$&$33.7$&$50.2$&$53.9$&$52.1$&$55.1$&$49.8$&$51.6$&$49.8$&$54.0$&$55.8$&$47.2$&$47.3$&$49.4$&$45.5$\\

$\text{\ w/\ }{\text{TokAlign}}$
&$48.1$&$35.7$&$34.7$&$32.8$&$32.9$&$33.4$&$32.8$&$52.1$&$\textbf{54.4}$&$53.4$&$55.2$&$50.3$&$52.4$&$50.4$&$54.8$&$59.8$&$47.5$&$47.4$&$50.2$&$46.2$\\

$\text{\ w/\ }{\text{TokAlign++}}$
&$49.9$&$37.4$&$33.9$&$34.0$&$34.2$&$35.1$&$34.7$&$\textbf{55.3}$&$50.4$&$54.8$&$\textbf{55.3}$&$\textbf{55.4}$&$53.4$&$52.2$&$\textbf{55.4}$&$61.0$&$48.1$&$47.9$&$50.6$&$47.3$\\[0.25em]\hdashline\\[-0.5em]

\multicolumn{21}{c}{\textit{Performance after 1k steps tuning on multilingual corpus.}}\\[0.3em]\hdashline\\[-0.4em]

$\text{\ w/\ }{\text{Focus}}$&$44.4$&$36.7$&$35.3$&$34.5$&$33.9$&$34.6$&$34.5$&$50.2$&$50.3$&$50.8$&$50.0$&$54.4$&$53.8$&$52.2$&$54.2$&$61.0$&$48.3$&$47.1$&$\textbf{54.2}$&$46.3$\\

$\text{\ w/\ }{\text{ZeTT}}$&$48.0$&$39.8$&$38.7$&$35.6$&$34.8$&$37.2$&$34.7$&$51.0$&$50.2$&$49.4$&$54.8$&$55.2$&$54.8$&$53.6$&$52.4$&$62.5$&$49.4$&$47.5$&$53.1$&$47.5$\\

$\text{\ w/\ }{\text{TokAlign}}$&$49.1$&$38.6$&$39.4$&$36.1$&$34.9$&$38.5$&$34.3$&$51.5$&$52.1$&$55.7$&$53.4$&$55.3$&$55.6$&$52.8$&$52.8$&$63.4$&$49.8$&$47.6$&$53.2$&$48.1$\\

$\text{\ w/\ }{\text{TokAlign++}}$&$50.8$&$\textbf{40.6}$&$\textbf{43.2}$&$\textbf{38.4}$&$\textbf{37.6}$&$\textbf{39.4}$&$\textbf{35.6}$&$54.5$&$52.7$&$\textbf{55.9}$&$54.9$&$55.3$&$\textbf{56.2}$&$\textbf{54.2}$&$54.4$&$\textbf{64.6}$&$\textbf{51.4}$&$\textbf{48.0}$&$53.5$&$\textbf{49.5}$\\

\midrule[0.8pt]

 $\text{Pythia}_{\text{6.9B}}$
&$54.4$&$\textbf{39.0}$&$\textbf{46.2}$&$39.3$&$39.8$&$39.3$&$\textbf{36.4}$&$43.8$&$40.2$&$50.2$&$54.2$&$50.2$&$\textbf{56.2}$&$54.4$&$52.2$&$70.4$&$53.9$&$\textbf{50.3}$&$53.8$&$48.6$\\[0.25em]\hdashline\\[-0.5em]

\multicolumn{21}{c}{\textit{Performance after initialization without any tuning.}} \\[0.3em]\hdashline\\[-0.4em]
 

$\text{\ w/\ }{\text{Focus}}$
&$33.1$&$32.7$&$33.7$&$33.4$&$33.3$&$33.0$&$32.1$&$54.4$&$\textbf{51.4}$&$49.1$&$\textbf{55.2}$&$50.9$&$53.8$&$51.0$&$53.8$&$45.5$&$46.7$&$45.9$&$50.1$&$44.2$\\

$\text{\ w/\ }{\text{ZeTT}}$
&$53.2$&$35.8$&$34.1$&$34.6$&$33.4$&$33.9$&$32.3$&$\textbf{55.3}$&$43.7$&$47.7$&$54.9$&$51.3$&$53.2$&$50.2$&$54.4$&$64.3$&$47.3$&$47.6$&$49.4$&$46.1$\\

$\text{\ w/\ }{\text{TokAlign}}$
&$52.6$&$35.1$&$34.4$&$34.3$&$34.5$&$33.8$&$33.2$&$54.5$&$47.8$&$55.1$&$53.1$&$52.9$&$53.8$&$51.8$&$\textbf{57.2}$&$64.7$&$47.7$&$47.2$&$50.2$&$47.0$\\

$\text{\ w/\ }{\text{TokAlign++}}$
&$53.2$&$36.7$&$35.0$&$34.7$&$34.6$&$35.5$&$34.4$&$55.1$&$50.4$&$\textbf{55.9}$&$\textbf{55.2}$&$55.2$&$53.4$&$53.4$&$54.6$&$66.5$&$48.4$&$46.5$&$50.2$&$47.8$\\[0.25em]\hdashline\\[-0.5em]

\multicolumn{21}{c}{\textit{Performance after 1k steps tuning on multilingual corpus.}} \\[0.3em]\hdashline\\[-0.4em]

$\text{\ w/\ }{\text{Focus}}$&$47.0$&$35.1$&$36.6$&$35.1$&$34.6$&$35.4$&$33.7$&$48.8$&$46.0$&$51.4$&$52.2$&$54.0$&$53.8$&$54.4$&$52.6$&$66.8$&$48.7$&$47.4$&$53.9$&$46.7$\\

$\text{\ w/\ }{\text{ZeTT}}$&$50.6$&$36.4$&$38.7$&$37.4$&$35.3$&$38.5$&$34.3$&$48.6$&$45.2$&$54.5$&$54.8$&$54.7$&$55.0$&$54.0$&$53.8$&$70.2$&$50.4$&$47.7$&$53.9$&$48.1$\\

$\text{\ w/\ }{\text{TokAlign}}$&$53.7$&$34.9$&$41.7$&$39.1$&$39.2$&$40.1$&$36.1$&$47.8$&$46.8$&$\textbf{55.9}$&$55.1$&$53.1$&$54.4$&$56.2$&$54.0$&$\textbf{71.5}$&$54.5$&$48.4$&$54.2$&$49.3$\\

$\text{\ w/\ }{\text{TokAlign++}}$&$\textbf{54.6}$&$36.7$&$42.7$&$\textbf{40.4}$&$\textbf{40.1}$&$\textbf{42.9}$&$36.3$&$51.0$&$48.7$&$\textbf{55.9}$&$55.0$&$\textbf{55.3}$&$55.2$&$\textbf{57.4}$&$54.8$&$\textbf{71.5}$&$\textbf{55.1}$&$48.2$&$\textbf{54.4}$&$\textbf{50.3}$\\

\bottomrule[1.2pt]
\end{tabular}
\end{table*}

Table \ref{tab:multilingual_ppl} shows the perplexity of initialized models and the ones after 1k language adaptation tuning steps. 
It can be found that the averaged perplexity of Pythia${}_{\text{1B}}$ and Pythia${}_{\text{6.9B}}$ after initialization using TokAlign++ (76.3) significantly outperforms the other three strong baseline methods, Focus (2.7$e^5$), ZeTT (1.7$e^2$), and our conference method, TokAlign (1.2$e^2$). 
After only 1k-step language adaptation tuning, TokAlign++ achieves an average improvement of 19.2\% compared to the vanilla model, whereas ZeTT still underperforms. 
It is observed that Pythia with TokAlign++ even achieves better performance than the Qwen2 models with a comparable parameter size on five low-resource languages. 
Across fifteen languages, the average token length after tokenization is 47.6\% shorter after switching to the new tokenizer. 

We further report the zero-shot in-context learning performance on four multilingual tasks in Table \ref{tab:multilingual_0_shot}. 
Models initialized by TokAlign++ are better than those using other methods like Focus (+3.9\%). 
The performance of other languages like Japanese (ja, +3.4\%) and Vietnamese (vi, +2.5\%) benefits from cross-lingual transfer. 
It is interesting to find that Pythia${}_{\text{1B}}$ initialized with TokAlign++ has a perplexity of 78.4, while its in-context learning performance is better than the 1k-step tuning model initialized by Focus. 
We attribute this to the English ability largely preserved by TokAlign++ (53.8\%), which significantly outperforms that of Focus (43.0\%). 

\paragraph{Cross-model Transferring Experiments} Unifying vocabulary with capable LLMs enables token-level distillation, which transfers their knowledge to smaller models and reduces the inference cost. 
We use training samples of the downstream tasks and the Pile corpus in the token-level distillation experiments. 
The output token predicted probabilities from the teacher model serve as the soft learning objective for Pythia. 
Specifically, we add the KL-divergence loss between the next token distribution from the teacher and student models to the original next token prediction loss for each distillation sample. 
In order to avoid a significant degradation in language modeling performance, we empirically set the proportion of training samples to 15\% following \citet{wei2023skywork}. 
There are two baseline methods used for comparison: ``+ Direct tuning'', in which models are fine-tuned directly on the training data, and ``+ Sentence distill'', where models are fine-tuned on the generated text from the teacher model given the question as a prompt. 

\begin{table*}[thp]

\renewcommand\arraystretch{1}

\centering
\scriptsize

\setlength{\tabcolsep}{0.7mm}
\caption{\label{tab:token_distill_res} The main results of token-level distillation on six downstream tasks with only 235M tokens. ``+Sentence distill'' denotes the sentence-level distillation results with Qwen2${}_{\text{7B}}$\cite{yang2024qwen2}, which fine-tunes on the output from Qwen2${}_{\text{7B}}$ given questions as prompt. ``0'' and ``5'' indicate the zero-shot and five-shot in-context learning results, respectively. 
}
 \begin{tabu}{lcccccccccccccc}
 
 \toprule[1.2pt]
  
  \multicolumn{1}{c}{ } & \multicolumn{2}{c}{\textbf{ARC-E}} & \multicolumn{2}{c}{\textbf{BoolQ}} & \multicolumn{2}{c}{\textbf{HellaSwag}} & \multicolumn{2}{c}{\textbf{OpenbookQA}} & \multicolumn{2}{c}{\textbf{PIQA}}  & \multicolumn{2}{c}{\textbf{WinoGrande}} & \multicolumn{2}{c}{\textbf{Avg}} \\

  \cmidrule(r){2-3} \cmidrule(r){4-5} \cmidrule(r){6-7} \cmidrule(r){8-9} \cmidrule(r){10-11} \cmidrule(r){12-13} \cmidrule(r){14-15} \noalign{\smallskip}

 \multicolumn{1}{c}{\textbf{Model}} & \textbf{0} & \textbf{5} & \textbf{0} & \textbf{5} & \textbf{0} & \textbf{5} & \textbf{0} & \textbf{5} & \textbf{0} & \textbf{5} & \textbf{0} & \textbf{5} & \textbf{0} &  \multicolumn{1}{c}{\textbf{5}} \\

\midrule[0.8pt]

$\text{Pythia}_{\text{1B}}$
&$56.82$&$58.71$&$60.43$&$57.37$&$37.68$&$37.66$&$18.80$&$19.00$&$70.40$&$71.49$&$53.20$&$52.01$&$49.55$&$49.37$\\

$\text{\ \ \ +\ Direct tuning}$
&$57.49$&$55.64$&$70.70$&$72.11$&$41.24$&$41.60$&$25.40$&$28.40$&$69.04$&$70.08$&$54.70$&$54.78$&$53.10$&$53.77$\\

$\text{\ \ \ +\ Sentence distill}$
&$52.27$&$53.41$&$67.49$&$67.06$&$39.03$&$39.08$&$21.80$&$22.80$&$66.97$&$68.99$&$51.85$&$52.17$&$49.90$&$50.58$\\[0.25em]\hdashline\\[-0.4em]

   
$\text{\ \ \ w/\ }{\text{Gemma}_{\text{7B}}}$
&$55.39$&$56.99$&$67.19$&$69.69$&$36.53$&$37.26$&$19.00$&$22.80$&$68.82$&$69.21$&$52.33$&$53.51$&$49.88$&$51.58$\\

$\text{\ \ \ w/\ }{\text{Qwen2}_{\text{7B}}}$
&$62.33$&$63.17$&$70.18$&$72.54$&$41.58$&$42.21$&$22.00$&$\textbf{28.20}$&$\textbf{73.01}$&$73.18$&$55.01$&$55.56$&$54.02$&$55.81$\\

$\text{\ \ \ w/\ }{\text{LLaMA3}_{\text{8B}}}$
&$\textbf{64.02}$&$\textbf{64.56}$&$\textbf{73.91}$&$\textbf{74.19}$&$\textbf{42.11}$&$\textbf{42.34}$&$\textbf{24.20}$&$27.60$&$72.74$&$\textbf{73.83}$&$\textbf{55.49}$&$\textbf{56.43}$&$\textbf{55.41}$&$\textbf{56.49}$\\

\midrule[0.8pt]

 $\text{Pythia}_{\text{6.9B}}$
&$65.99$&$69.23$&$62.84$&$62.02$&$47.56$&$47.64$&$25.00$&$27.00$&$74.65$&$75.41$&$60.46$&$62.43$&$56.08$&$57.29$\\

$\text{\ \ \ +\ Direct tuning}$
&$66.25$&$66.20$&$79.30$&$78.87$&$52.21$&$53.39$&$33.20$&$33.00$&$72.91$&$74.48$&$62.90$&$61.72$&$61.13$&$61.28$\\

$\text{\ \ \ +\ Sentence distill}$
&$61.70$&$65.36$&$76.64$&$76.88$&$48.98$&$51.33$&$28.20$&$30.40$&$70.18$&$71.55$&$58.96$&$62.19$&$57.44$&$59.62$\\[0.25em]\hdashline\\[-0.4em]


$\text{\ \ \ w/\ }{\text{Gemma}_{\text{7B}}}$
&$67.59$&$68.94$&$76.06$&$75.66$&$47.83$&$48.36$&$28.40$&$31.40$&$73.78$&$75.52$&$59.04$&$64.17$&$58.78$&$60.67$\\

$\text{\ \ \ w/\ }{\text{Qwen2}_{\text{7B}}}$
&$\textbf{71.72}$&$\textbf{73.27}$&$\textbf{79.85}$&$\textbf{80.00}$&$\textbf{50.78}$&$\textbf{51.12}$&$\textbf{29.20}$&$\textbf{34.00}$&$\textbf{77.26}$&$\textbf{77.91}$&$\textbf{61.33}$&$\textbf{64.56}$&$\textbf{61.69}$&$\textbf{63.48}$\\

$\text{\ \ \ w/\ }{\text{LLaMA3}_{\text{8B}}}$
&$67.05$&$69.78$&$77.83$&$78.78$&$48.83$&$50.15$&$26.00$&$32.00$&$74.21$&$76.22$&$60.22$&$60.93$&$59.02$&$61.31$\\

\bottomrule[1.2pt]
\end{tabu}

\end{table*}

Table \ref{tab:token_distill_res} presents the results of two baseline methods and token-level distillation from three teacher models under 235M tokens tuning budget. 
We can find that token-level distillation significantly outperforms the sentence-level distillation method. 
In the neural machine translation domain, token-level distillation is better than sentence-level distillation when employing larger student models, simpler textual content, and abundant decoding information~\cite{kim-rush-2016-sequence, wei2024sent}. 
Token-level distillation on Qwen2${}_{\text{7B}}$ is 4.4\% higher than sentence-level distillation on average. 
After token-level distillation, Pythia${}_{\text{1B}}$ even performs on par with the vanilla Pythia${}_{\text{7B}}$. 
Furthermore, it is noted that knowledge transfer among LLMs will be limited to sentence-level distillation without vocabulary unification, which highlights the significance of unifying tokenizers across large language models. 


\begin{table*}[thp]

\renewcommand\arraystretch{1.05}

\centering
\scriptsize

\setlength{\tabcolsep}{0.4mm}

\caption{\label{tab:main_res_baseline} The main results of replacing the vocabulary of Pythia or LLaMA3 to Gemma. The best performance among the eight methods is displayed in \textbf{bold}. 
}

 \begin{tabular}{lrcccccccccccccc}
 
 \toprule[1.2pt]
  
  \multicolumn{2}{c}{ } & \multicolumn{2}{c}{\textbf{ARC-E}} & \multicolumn{2}{c}{\textbf{BoolQ}} & \multicolumn{2}{c}{\textbf{HellaSwag}} & \multicolumn{2}{c}{\textbf{OpenbookQA}} & \multicolumn{2}{c}{\textbf{PIQA}}  & \multicolumn{2}{c}{\textbf{WinoGrande}} & \multicolumn{2}{c}{\textbf{Avg}} \\

  \cmidrule(r){3-4} \cmidrule(r){5-6} \cmidrule(r){7-8} \cmidrule(r){9-10} \cmidrule(r){11-12} \cmidrule(r){13-14} \cmidrule(r){15-16} \noalign{\smallskip}

 \multicolumn{1}{c}{\textbf{Model}} & \multicolumn{1}{c}{\tiny $\#$\textbf{GPU Hour}} & \textbf{0} & \textbf{5} & \textbf{0} & \textbf{5} & \textbf{0} & \textbf{5} & \textbf{0} & \textbf{5} & \textbf{0} & \textbf{5} & \textbf{0} & \textbf{5} & \textbf{0} &  \multicolumn{1}{c}{\textbf{5}} \\

\midrule[0.8pt]

$\text{Pythia}_{\text{1B}}$
&\multicolumn{1}{c}{$-$}&$56.82$&$58.71$&$60.43$&$57.37$&$37.68$&$37.66$&$18.80$&$19.00$&$70.40$&$71.49$&$53.20$&$52.01$&$49.55$&$49.37$\\[0.25em]\hdashline\\[-0.4em]


$\text{\ w$/$\ Rand. Init.}$
&$99.70\ $&$31.36$&$31.61$&$37.83$&$49.11$&$26.35$&$26.40$&$14.00$&$12.60$&$54.57$&$55.33$&$49.17$&$49.17$&$35.55$&$37.37$\\

$\text{\ w$/$\ Rand. Perm.}$
&$99.70\ $&$31.69$&$32.95$&$37.77$&$54.80$&$26.43$&$26.39$&$14.00$&$12.60$&$55.50$&$55.98$&$47.04$&$50.67$&$35.40$&$38.90$\\

$\text{\ w$/$\ Multivariate}$
&$99.70\ $&$32.79$&$34.18$&$45.08$&$49.72$&$27.67$&$27.87$&$15.20$&$16.20$&$56.09$&$57.83$&$50.51$&$50.12$&$37.89$&$39.32$\\

$\text{\ w$/$\ Mean}$
&$99.70\ $&$44.87$&$46.97$&$53.39$&$55.20$&$31.59$&$31.67$&$16.20$&$17.00$&$61.32$&$62.46$&$49.25$&$51.85$&$42.77$&$44.19$\\

$\text{\ w$/$\ OFA}$
&$99.70\ $&$38.17$&$37.79$&$55.14$&$52.35$&$28.29$&$28.62$&$14.40$&$12.20$&$58.43$&$58.54$&$49.96$&$50.99$&$40.73$&$40.08$\\

$\text{\ w$/$\ WECHSEL}$
&$99.70\ $&$43.35$&$45.33$&$56.61$&$54.34$&$32.53$&$32.41$&$14.80$&$16.20$&$61.70$&$62.89$&$52.01$&$52.72$&$43.50$&$43.98$\\

$\text{\ w$/$\ Focus}$
&$99.70\ $&$46.55$&$48.95$&$56.21$&$\textbf{55.78}$&$32.27$&$32.46$&$19.20$&$18.00$&$63.82$&$64.80$&$51.70$&$51.78$&$44.96$&$45.29$\\

$\text{\ w$/$\ ZeTT}$
&$418.94\ $&$47.14$&$49.03$&$57.06$&$53.70$&$34.06$&$34.06$&$18.40$&$19.40$&$64.15$&$65.34$&$52.09$&$51.22$&$45.48$&$45.46$\\

$\text{\ w$/$\ TokAlign}$
&$99.70\ $&$54.46$&$56.86$&$58.90$&$52.26$&$36.16$&$\textbf{36.27}$&$\textbf{21.00}$&$20.20$&$67.74$&$68.50$&$52.25$&$50.91$&$48.42$&$47.50$\\

$\text{\ w$/$\ TokAlign++}$
&$\textbf{19.94}\ $&$\textbf{55.64}$&$56.94$&$\textbf{60.55}$&$54.25$&$\textbf{36.33}$&$36.24$&$19.60$&$\textbf{21.40}$&$\textbf{68.34}$&$\textbf{69.48}$&$52.25$&$\textbf{53.20}$&$\textbf{48.78}$&$\textbf{48.58}$\\

$\text{\ \ \ \ \ w$/$\ Hid. Rep.}$
&$\textbf{19.94}\ $&$54.46$&$\textbf{57.07}$&$60.03$&$52.20$&$35.98$&$35.96$&$20.00$&$\textbf{21.40}$&$68.28$&$69.10$&$\textbf{52.57}$&$52.57$&$48.55$&$48.05$\\

\midrule[0.8pt]

 $\text{Pythia}_{\text{2.8B}}$
&\multicolumn{1}{c}{$-$}&$63.80$&$67.00$&$63.91$&$65.14$&$45.32$&$45.04$&$24.00$&$25.20$&$74.05$&$74.43$&$58.64$&$60.77$&$54.95$&$56.26$\\[0.25em]\hdashline\\[-0.4em]


$\text{\ w$/$\ Rand. Init.}$
&$194.78\ $&$30.47$&$32.91$&$38.20$&$51.07$&$26.46$&$26.69$&$14.40$&$13.20$&$55.17$&$55.06$&$48.30$&$50.51$&$35.50$&$38.24$\\

$\text{\ w$/$\ Rand. Perm.}$
&$194.78\ $&$31.48$&$31.86$&$37.83$&$50.46$&$26.48$&$26.49$&$13.60$&$14.40$&$54.03$&$54.95$&$50.20$&$48.86$&$35.60$&$37.84$\\

$\text{\ w$/$\ OFA}$
&$194.78\ $&$50.13$&$54.12$&$60.89$&$61.47$&$36.39$&$36.88$&$18.00$&$19.00$&$65.18$&$64.80$&$54.06$&$54.85$&$47.44$&$48.52$\\

$\text{\ w$/$\ WECHSEL}$
&$194.78\ $&$52.48$&$54.92$&$59.42$&$56.76$&$36.79$&$37.30$&$19.20$&$20.80$&$64.04$&$64.25$&$56.43$&$55.72$&$48.06$&$48.29$\\

$\text{\ w$/$\ Focus}$
&$194.78\ $&$54.29$&$58.16$&$61.44$&$62.84$&$38.38$&$39.09$&$20.00$&$20.20$&$68.44$&$68.28$&$54.62$&$56.04$&$49.53$&$50.77$\\

$\text{\ w$/$\ ZeTT}$
&$855.96\ $&$57.15$&$59.42$&$61.68$&$62.05$&$42.17$&$42.25$&$21.80$&$23.60$&$71.11$&$71.16$&$56.59$&$59.19$&$51.75$&$52.95$\\

$\text{\ w$/$\ TokAlign}$
&$194.78\ $&$61.62$&$65.15$&$63.82$&$\textbf{65.47}$&$43.13$&$43.18$&$23.40$&$\textbf{25.80}$&$\textbf{72.14}$&$72.42$&$\textbf{58.17}$&$\textbf{61.17}$&$53.71$&$\textbf{55.53}$\\

$\text{\ w$/$\ TokAlign++}$
&$\textbf{38.96}\ $&$\textbf{61.91}$&$\textbf{66.04}$&$\textbf{64.43}$&$60.98$&$\textbf{43.42}$&$\textbf{43.42}$&$\textbf{24.80}$&$\textbf{25.80}$&$\textbf{72.14}$&$\textbf{72.96}$&$56.75$&$61.09$&$\textbf{53.91}$&$55.05$\\

\midrule[0.8pt]

 $\text{LLaMA3}_{\text{8B}}$
&\multicolumn{1}{c}{$-$}&$80.22$&$83.92$&$81.13$&$82.17$&$60.16$&$61.58$&$34.40$&$36.60$&$79.43$&$80.41$&$73.24$&$77.43$&$68.10$&$70.35$\\[0.25em]\hdashline\\[-0.4em]

$\text{\ w$/$\ WECHSEL}$
&$986.03\ $&$65.19$&$68.31$&$74.43$&$76.97$&$52.42$&$53.62$&$26.60$&$30.40$&$73.78$&$74.27$&$68.98$&$70.48$&$60.23$&$62.34$\\

$\text{\ w$/$\ Focus}$
&$986.03\ $&$67.30$&$69.57$&$78.99$&$77.92$&$53.56$&$54.78$&$28.20$&$31.20$&$74.59$&$75.19$&$70.01$&$73.64$&$62.11$&$63.72$\\

$\text{\ w$/$\ ZeTT}$
&$2297.14\ $&$73.86$&$75.17$&$78.35$&$78.38$&$54.08$&$54.93$&$27.40$&$31.60$&$75.90$&$76.66$&$70.01$&$71.82$&$63.27$&$64.76$\\

$\text{\ w$/$\ TokAlign}$
&$986.03\ $&$73.11$&$76.68$&$75.78$&$80.40$&$54.79$&$55.77$&$29.80$&$32.40$&$76.06$&$77.64$&$\textbf{71.35}$&$74.90$&$63.48$&$66.30$\\

$\text{\ w$/$\ TokAlign++}$
&$\textbf{197.21}\ $&$\textbf{75.46}$&$\textbf{78.03}$&$\textbf{80.61}$&$\textbf{82.60}$&$\textbf{57.55}$&$\textbf{58.51}$&$\textbf{31.20}$&$\textbf{33.80}$&$\textbf{77.80}$&$\textbf{78.67}$&$71.19$&$\textbf{75.85}$&$\textbf{65.64}$&$\textbf{67.91}$\\

\bottomrule[1.2pt]
\end{tabular}
\end{table*}

\begin{figure}[b]
    \centering
    \includegraphics [scale=0.46]{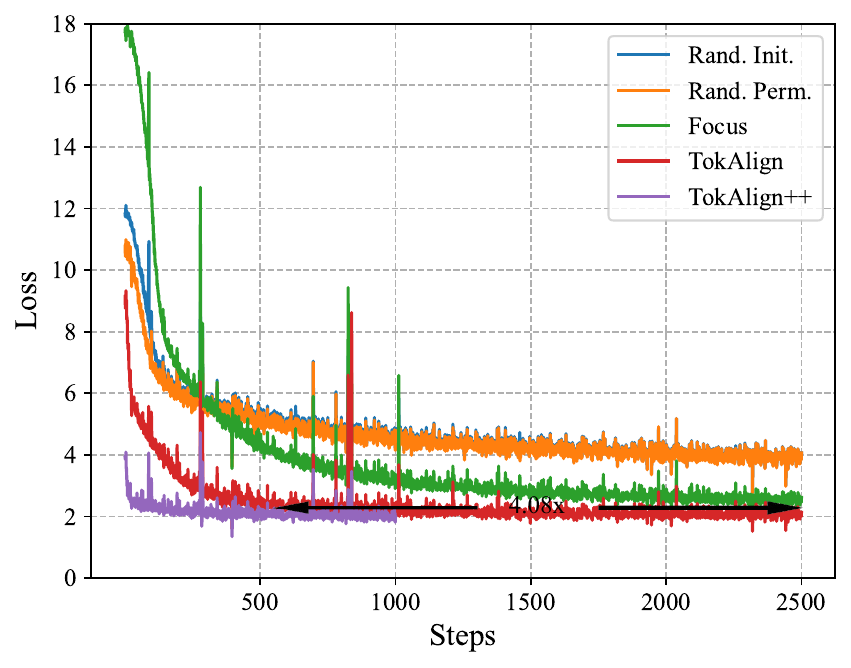}
    \caption{\label{fig:2b8_loss}The training loss curve of Pythia${}_{\text{2.8B}}$.}
\end{figure}

\paragraph{Comparison with Other Vocabulary Adaptation Methods} 
We report experimental results of all methods when replacing the Pythia vocabulary (50.3k) with the Gemma vocabulary (256.0k) in Table \ref{tab:main_res_baseline}. 
It can be found that TokAlign++ outperforms baseline methods and our conference method TokAlign~\cite{li-etal-2025-tokalign} using only 25\% tuning steps of other methods. 
TokenAlign++ achieves a 2.7\% performance improvement on average over the strong baseline ZeTT. 
After new vocabulary adaptation, TokAlign++ still preserves 98.2\% of vanilla model capabilities. 
ZeTT needs more computational resources to train a hypernetwork for parameter prediction, such as 661.2 GPU hours for the Pythia${}_{\text{2.8B}}$ model, whereas our approach requires under two hours on a 128-core CPU server to train GloVe embeddings and conduct token alignment. 
Figure \ref{fig:2b8_loss} shows loss curves of Pythia${}_{\text{2.8B}}$ using different methods, where TokAlign++ brings the lowest first-step training loss and the fastest convergence compared to other baseline methods. 
It reaches 2.75 at the 612th training step and is 4.08$\times$ (2500/612) speed up than Focus. 
We further propose a method to extract token representations from the internal hidden states of the vanilla language model, reducing the cost to train GloVe embeddings. 
As shown in the ``w/ Hid. Rep.'' row of Table \ref{tab:main_res_baseline}, it achieves in-context learning results comparable with the GloVe embeddings for token alignment.

\begin{table*}[thp]

\renewcommand\arraystretch{1.05}

\centering
\scriptsize

\setlength{\tabcolsep}{0.6mm}
\caption{\label{tab:main_res} The benchmark results of replacing different tokenizers using TokAlign++. The overlapping ratios between the vocabulary of Pythia and other models are 6.23\% (Gemma), 26.92\% (Qwen2), 28.10\% (LLaMA2), and 32.85\% (LLaMA3).  
}
 \begin{tabular}{lrcccccccccccccc}
 
 \toprule[1.2pt]
  
  \multicolumn{2}{c}{ } & \multicolumn{2}{c}{\textbf{ARC-E}} & \multicolumn{2}{c}{\textbf{BoolQ}} & \multicolumn{2}{c}{\textbf{HellaSwag}} & \multicolumn{2}{c}{\textbf{OpenbookQA}} & \multicolumn{2}{c}{\textbf{PIQA}}  & \multicolumn{2}{c}{\textbf{WinoGrande}} & \multicolumn{2}{c}{\textbf{Avg}} \\

  \cmidrule(r){3-4} \cmidrule(r){5-6} \cmidrule(r){6-8} \cmidrule(r){9-10} \cmidrule(r){11-12} \cmidrule(r){13-14} \cmidrule(r){15-16} \noalign{\smallskip}

 \multicolumn{1}{c}{\textbf{Model}} & \multicolumn{1}{c}{$\#$\textbf{$\mathcal{V}$ (k)}} & \textbf{0} & \textbf{5} & \textbf{0} & \textbf{5} & \textbf{0} & \textbf{5} & \textbf{0} & \textbf{5} & \textbf{0} & \textbf{5} & \textbf{0} & \textbf{5} & \textbf{0} &  \multicolumn{1}{c}{\textbf{5}} \\

\midrule[0.8pt]

$\text{Pythia}_{\text{1B}}$
&$50.3$&$56.82$&$58.71$&$60.43$&$57.37$&$37.68$&$37.66$&$18.80$&$19.00$&$70.40$&$71.49$&$53.20$&$52.01$&$49.55$&$49.37$\\[0.25em]\hdashline\\[-0.4em]


$\text{\ \ \ $\to$\ }{\text{Gemma}}$
&$256.0$&$\textbf{55.64}$&$56.94$&$\textbf{60.55}$&$54.25$&$36.33$&$36.24$&$19.60$&$21.40$&$68.34$&$69.48$&$52.25$&$53.20$&$48.78$&$48.58$\\

$\text{\ \ \ $\to$\ }{\text{Qwen2}}$
&$152.1$&$54.25$&$57.37$&$59.36$&$55.11$&$\textbf{36.76}$&$\textbf{36.86}$&$\textbf{20.80}$&$21.00$&$68.34$&$69.26$&$53.75$&$53.12$&$48.87$&$48.78$\\

$\text{\ \ \ $\to$\ }{\text{LLaMA2}}$
&$32.0$&$49.79$&$53.11$&$60.31$&$\textbf{56.91}$&$35.70$&$35.70$&$18.40$&$19.40$&$66.59$&$66.65$&$\textbf{53.91}$&$\textbf{54.54}$&$47.45$&$47.72$\\

$\text{\ \ \ $\to$\ }{\text{LLaMA3}}$
&$128.0$&$55.30$&$\textbf{57.95}$&$59.33$&$54.34$&$36.66$&$36.77$&$20.60$&$\textbf{21.60}$&$\textbf{68.88}$&$\textbf{69.70}$&$52.64$&$53.59$&$\textbf{48.90}$&$\textbf{48.99}$\\

\midrule[0.8pt]

 $\text{Pythia}_{\text{2.8B}}$
&$50.3$&$63.80$&$67.00$&$63.91$&$65.14$&$45.32$&$45.04$&$24.00$&$25.20$&$74.05$&$74.43$&$58.64$&$60.77$&$54.95$&$56.26$\\[0.25em]\hdashline\\[-0.4em]


$\text{\ \ \ $\to$\ }{\text{Gemma}}$
&$256.0$&$61.91$&$\textbf{66.04}$&$\textbf{64.43}$&$60.98$&$43.42$&$43.42$&$\textbf{24.80}$&$25.80$&$72.14$&$72.96$&$56.75$&$\textbf{61.09}$&$53.91$&$55.05$\\

$\text{\ \ \ $\to$\ }{\text{Qwen2}}$
&$152.1$&$\textbf{62.63}$&$64.94$&$63.27$&$\textbf{63.49}$&$\textbf{44.10}$&$43.76$&$23.60$&$\textbf{26.80}$&$73.45$&$73.18$&$58.33$&$58.72$&$54.23$&$55.15$\\

$\text{\ \ \ $\to$\ }{\text{LLaMA3}}$
&$128.0$&$62.58$&$65.45$&$63.33$&$62.91$&$44.03$&$\textbf{44.18}$&$23.60$&$26.20$&$\textbf{73.56}$&$\textbf{73.72}$&$\textbf{59.35}$&$60.30$&$\textbf{54.83}$&$\textbf{55.46}$\\

\midrule[0.8pt]

 $\text{Pythia}_{\text{6.9B}}$
&$50.3$&$65.99$&$69.23$&$62.84$&$62.02$&$47.56$&$47.64$&$25.00$&$27.00$&$74.65$&$75.41$&$60.46$&$62.43$&$56.08$&$57.29$\\[0.25em]\hdashline\\[-0.4em]


$\text{\ \ \ $\to$\ }{\text{Gemma}}$
&$256.0$&$65.66$&$67.34$&$63.46$&$61.90$&$45.75$&$46.01$&$23.40$&$26.80$&$73.67$&$73.56$&$\textbf{60.93}$&$61.96$&$55.48$&$56.26$\\

$\text{\ \ \ $\to$\ }{\text{Qwen2}}$
&$152.1$&$66.16$&$68.01$&$65.02$&$\textbf{62.66}$&$\textbf{46.88}$&$\textbf{46.80}$&$\textbf{24.80}$&$27.60$&$72.74$&$73.39$&$60.46$&$62.67$&$56.01$&$56.86$\\

$\text{\ \ \ $\to$\ }{\text{LLaMA3}}$
&$128.0$&$\textbf{66.71}$&$\textbf{69.23}$&$\textbf{65.14}$&$60.55$&$46.84$&$46.73$&$\textbf{24.80}$&$\textbf{28.60}$&$\textbf{73.83}$&$\textbf{75.08}$&$60.46$&$\textbf{62.98}$&$\textbf{56.30}$&$\textbf{57.20}$\\

\bottomrule[1.2pt]
\end{tabular}
\end{table*}

\paragraph{Generalize to More LLMs} There are some open-weight large language models that do not publicize their pre-training corpora, like LLaMA3~\cite{meta2024llama3}. 
It is more challenging for vocabulary adaptation methods to replace their vocabulary while keeping the original performance than fully open-source LLMs. 
Table \ref{tab:main_res_baseline} reports the results of replacing the LLaMA3 tokenizer with the more efficient Gemma tokenizer. 
We can find that TokAlign++ outperforms other baseline methods with much less GPU hours cost, recovering 96.46\% performance on average. 
The improvement over our conference method TokAlign reaches 1.88\% across six in-context learning benchmarks. 

\begin{figure}[b]
    \centering
    \includegraphics [scale=0.46]{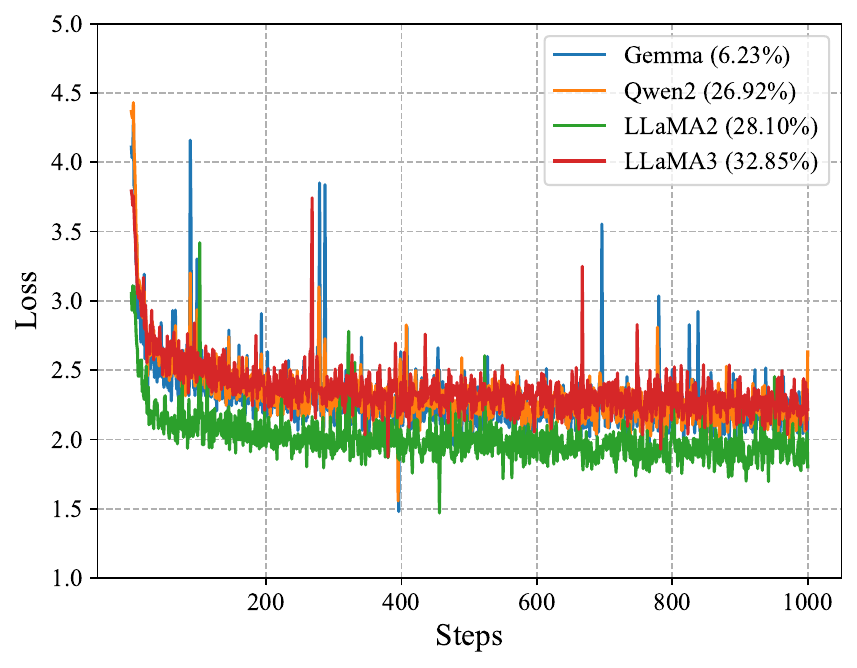}
    \vspace{-2mm}
    \caption{\label{fig:1b2other_loss}The training loss curve of Pythia${}_{\text{1B}}$ for different overlapping ratios.}
\end{figure}

\paragraph{Generalize to Different Tokenizers} TokAlign++ is also used to replace the following three target tokenizers: LLaMA2, LLaMA3, and Qwen2. 
We show the in-context learning performance of Pythia after vocabulary adaptation across three parameter amounts in Table \ref{tab:main_res}. 
With only 1k steps, TokAlign++ recovers an average of 98.6\% of original performance. 
Given the target vocabulary with more than 50.3k tokens, which is larger than the amount of tokens in Pythia vocabulary, our method performs better when there is a higher overlapping ratio between source and target vocabularies, ranging from 98.31\% for Gemma to 99.43\% for LLaMA3. 
After switching to the LLaMA3 tokenizer, the zero-shot in-context learning performance of Pythia${}_{\text{6.9B}}$ even exceeds the vanilla base model. 
The result for LLaMA2 vocabulary is only 96.2\%, falling below the average performance. 
This decline may be attributed to the loss of 75M embedding parameters (7.4\% of Pythia${}_{\text{1B}}$) when reducing the vocabulary size from 50.3k to 32.0k. 

Figure \ref{fig:1b2other_loss} illustrates the training loss curve of Pythia${}_{\text{1B}}$ using TokAlign++. 
The convergence process for these settings is similar, which differs from the finding in our conference paper that lower overlapping brings a slower convergence process. 
We argue that this may come from the significantly better initialized parameters, reducing the first step training loss from 9.5 to 4.1 when changing to the Gemma tokenizer. 

\section{Experimental Analysis}
\label{sec:analysis}
\subsection{Alignment Evaluation Metrics}

\begin{figure*}[b]
    \centering
    \subfigure[BLEU-1($\mathcal{C}_{s}$, $\mathcal{C}_{s}^{'}$)]{\includegraphics [scale=0.36]{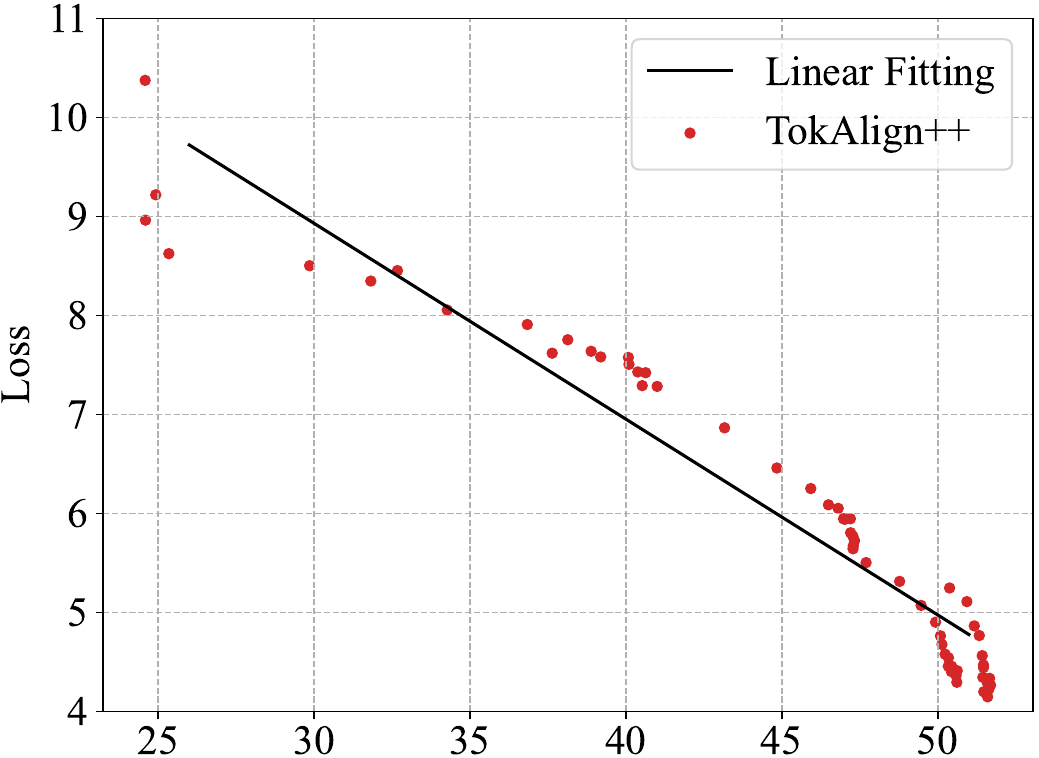}\label{fig:bleu1_loss}}
    \subfigure[BERTScore($\mathcal{C}$, $\mathcal{C}^{'}$) for $M_{t\to s}$]{\includegraphics [scale=0.36]{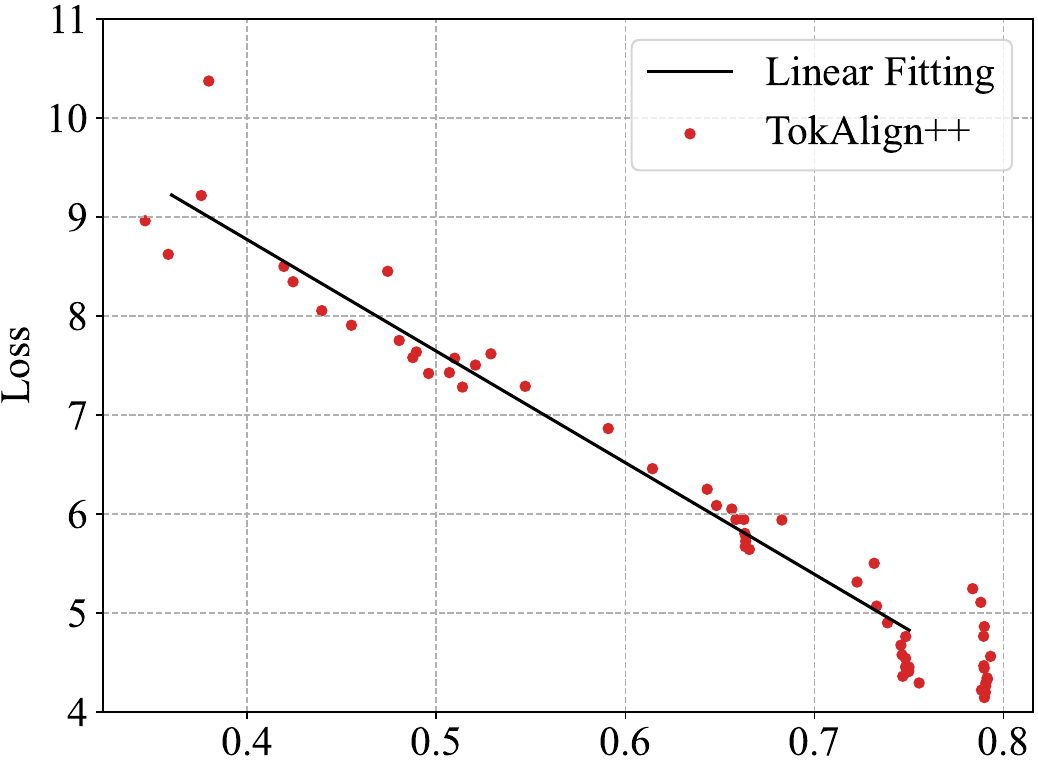}\label{fig:bertscore_loss}}
    \subfigure[BLEU-1($\mathcal{C}_{t}$, $\mathcal{C}_{t}^{'}$)]{\includegraphics [scale=0.36]{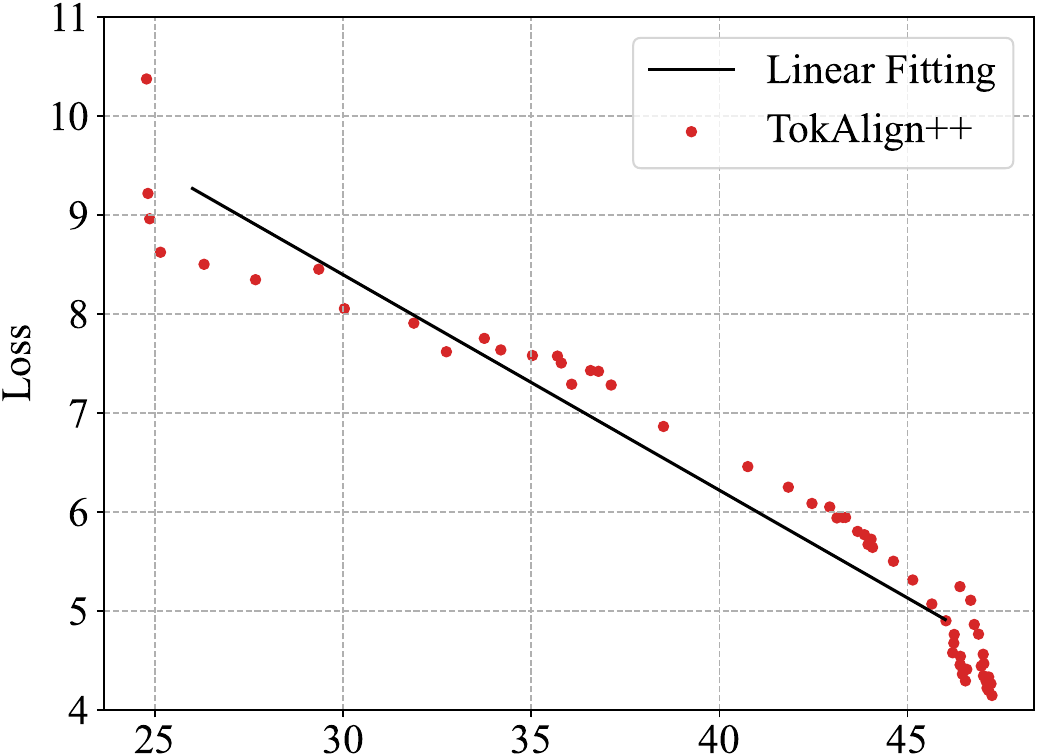}\label{fig:bleu1_loss_target}}
    \subfigure[BERTScore($\mathcal{C}$, $\mathcal{C}^{'}$) for $M_{s\to t}$]{\includegraphics [scale=0.36]{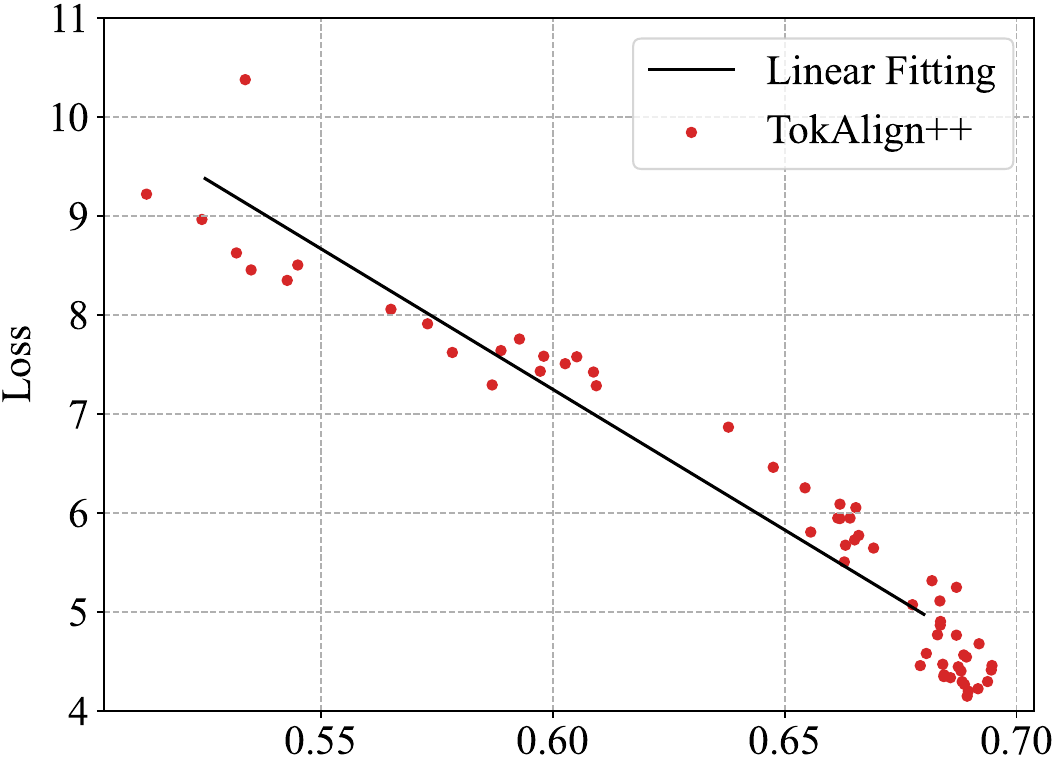}\label{fig:bertscore_loss_target}}
    \caption{\label{fig:loss_fit}The relationship between initial training loss and BLEU-1 (a, c) or BERTScore (b, d) for Pythia$_{\text{1B}}$.}
\end{figure*}

The impact of the learned alignment matrix $M_{t\to s}$ is further investigated by changing the hyper-parameters of GloVe. 
Given the same evaluation corpus, different alignment matrices $M_{t\to s}$ yield varying initial parameters and result in distinct BLEU-1 or BERTScore scores. 
The relationship between BLEU-1 and the first-step training loss is shown in Figure \ref{fig:bleu1_loss}. 
We can find that there is a clear negative relationship, where the correlation coefficient $r^2$ reaches 0.93. 
In the BERTScore evaluation, the sentence embedding model named ``all-mpnet-base-v2''~\cite{song2020mpnet} is chosen for its superior quality on the sentence embedding benchmark. 
As shown in Figure \ref{fig:bertscore_loss}, this metric also exhibits a clear negative relationship with the first step training loss ($r^2$=0.95). 

On the other hand, we can evaluate the token alignment from the source token to the target token direction. 
The token alignment matrix $M_{s\to t}$ is applied to the tokenized corpus $\mathcal{C}_{s}$, resulting in the converted target token corpus $\mathcal{C}_{t}^{'}$. 
Then BLEU-1($\mathcal{C}_{t}$, $\mathcal{C}_{t}^{'}$) is used to quantify the performance of token alignment. 
Figure \ref{fig:bleu1_loss_target} illustrates the negative relationship with the first step training loss, which is similar to Figure \ref{fig:bleu1_loss}. 
For the semantic similarity analysis, the recovered test corpus $\mathcal{C}^{'}$ is de-tokenzied from $\mathcal{C}_{t}^{'}$ using Tokenizer${}_{t}$. 
The result of BERTScore($\mathcal{C}$, $\mathcal{C}^{'}$) for $M_{s\to t}$ is shown in Figure \ref{fig:bertscore_loss_target}. 
We can also find a clear negative relationship like Figure \ref{fig:bertscore_loss} ($r^2$=0.93). 
Therefore, the BLEU-1 score and BERTScore for the alignment matrix $M_{t\to s}$ and $M_{s\to t}$ can serve as reliable quality indicators for the initial parameter: a higher score indicates a better initial parameter. 

\begin{figure}[b]
    \centering
    \subfigure[Vocabulary coverage]{\includegraphics [scale=0.46]{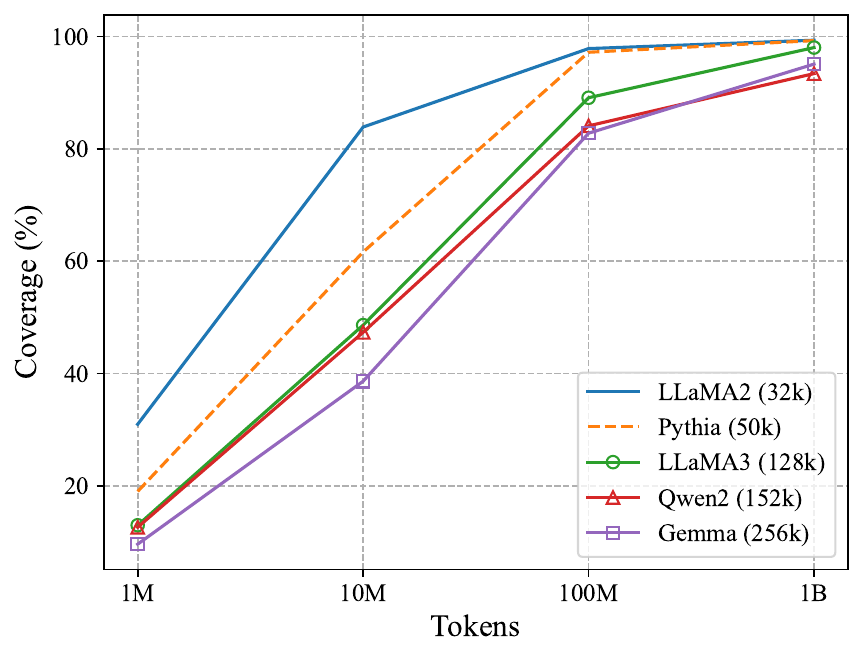}\label{fig:1b2glove_vec}}
    \subfigure[Initial loss with Gemma tokenizer]{\includegraphics [scale=0.46]{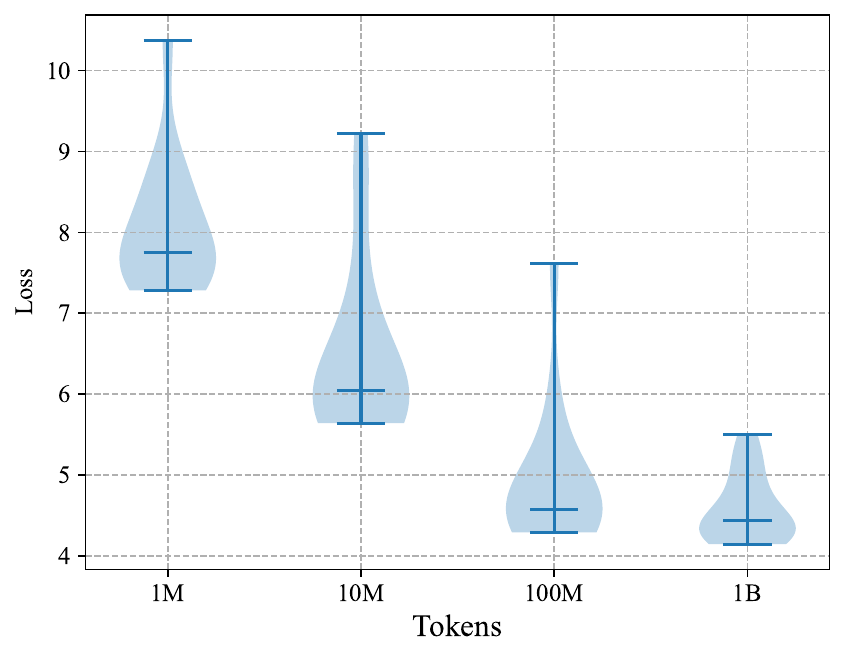}\label{fig:1b2glove_loss}}
    \vspace{-2mm}
    \caption{\label{fig:glove_hyper}The average vocabulary coverage (a) and initial training loss of Pythia${}_{\text{1B}}$ (b) when different amounts of tokens are used to train the GloVe vector.}
\end{figure}

\subsection{Ablation Study}
\label{sec:ablation_token_rep}
In this section, we provide extensive ablation studies to investigate the contribution of each component in TokAlign++.

\paragraph{Token Representation Analysis} The number of tokens used to train the GloVe vector will affect the quality of token representation, resulting in different token alignment results and initial parameters. 
Figure \ref{fig:1b2glove_vec} shows the vocabulary coverage under different train token budgets. 
We can find that the training corpus with 100M tokens contains more than 80\% tokens in the five vocabularies involved. 
As for the default 1B tokens budget, it covers more than 90\% tokens in the vocabulary. 
Figure \ref{fig:1b2glove_loss} illustrates the distribution of the first training loss given four training token budgets. 
The training loss of models initialized by TokAlign++ decreases with more training tokens used, and converges to 4 when the token representations are trained with 1B tokens. 
We further compare methods to train the token representations and report the results in Table \ref{tab:ablation_res}. 
Given the same corpus with 100M tokens, GloVe significantly outperforms the other two slide window methods on the training loss and six downstream benchmarks. 
Therefore, we adopt GloVe to train token representations for better vocabulary adaptation performance. 

\begin{table*}[thp]

\renewcommand\arraystretch{1.2}

\centering
\scriptsize

\setlength{\tabcolsep}{0.68mm}
\caption{\label{tab:ablation_res} The benchmark results of replacing Gemma Tokenizer for Pythia${}_{\text{1B}}$ using TokAlign++ without any tuning under different settings. ${}^*$ denotes the cross-scale generalization experiment, where the hidden states are extracted from Pythia${}_{\text{6.9B}}$.
}
 \begin{tabular}{lccccccccccccccr}
 
 \toprule[1.2pt]
  
  \multicolumn{1}{c}{ } & \multicolumn{2}{c}{\textbf{ARC-E}} & \multicolumn{2}{c}{\textbf{BoolQ}} & \multicolumn{2}{c}{\textbf{HellaSwag}} & \multicolumn{2}{c}{\textbf{OpenbookQA}} & \multicolumn{2}{c}{\textbf{PIQA}}  & \multicolumn{2}{c}{\textbf{WinoGrande}} & \multicolumn{2}{c}{\textbf{Avg $\uparrow$}} & \multicolumn{1}{c}{ }\\

  \cmidrule(r){2-3} \cmidrule(r){4-5} \cmidrule(r){6-7} \cmidrule(r){8-9} \cmidrule(r){10-11} \cmidrule(r){12-13} \cmidrule(r){14-15} \noalign{\smallskip}

 \multicolumn{1}{c}{\textbf{Setting}} & \textbf{0} & \textbf{5} & \textbf{0} & \textbf{5} & \textbf{0} & \textbf{5} & \textbf{0} & \textbf{5} & \textbf{0} & \textbf{5} & \textbf{0} & \textbf{5} & \textbf{0} &  \multicolumn{1}{c}{\textbf{5}} & \multicolumn{1}{c}{\textbf{Loss} $\downarrow$} \\

\midrule[0.8pt]\\[-1.2em]

\multicolumn{16}{c}{\textit{Performance using different methods to train token representations under 100M tokens budget.}} \\[0.5em]\hdashline\\[-0.4em]

$\text{CBOW}$
&$27.36$&$25.72$&$53.58$&$49.63$&$25.99$&$26.49$&$\textbf{19.40}$&$16.60$&$51.90$&$52.72$&$50.12$&$49.41$&$38.06$&$36.76$&$7.27$\\

$\text{Skip-Gram}$
&$36.87$&$38.43$&$59.14$&$52.81$&$29.34$&$29.36$&$15.80$&$16.00$&$58.05$&$57.18$&$49.25$&$50.28$&$41.41$&$40.68$&$4.70$\\

$\text{GloVe}$
&$44.23$&$46.46$&$\textbf{59.85}$&$53.82$&$31.62$&$31.38$&$18.80$&$17.60$&$60.94$&$60.83$&$50.91$&$53.35$&$44.39$&$43.91$&$4.36$\\[0.25em]\hdashline\\[-0.7em]

\multicolumn{16}{c}{\textit{Performance using different methods to obtain the token representation from hidden states of the last layer. }} \\[0.5em]\hdashline\\[-0.4em]

$\text{Max Pooling}$
&$30.77$&$30.68$&$37.95$&$53.82$&$26.04$&$26.05$&$14.80$&$15.20$&$55.11$&$55.11$&$45.94$&$48.78$&$35.10$&$38.27$&$8.79$\\

$\text{Avg. Pooling}$
&$35.73$&$33.88$&$51.31$&$58.53$&$26.55$&$26.61$&$17.00$&$16.20$&$56.64$&$55.77$&$49.80$&$49.57$&$39.51$&$40.09$&$8.34$\\

$\text{Last Token}$
&$39.27$&$37.75$&$49.30$&$\textbf{60.00}$&$28.56$&$28.23$&$16.20$&$16.60$&$59.63$&$57.67$&$50.59$&$51.85$&$40.59$&$42.02$&$6.03$\\

$\text{Last Token}^*$
&$36.49$&$36.36$&$37.83$&$53.91$&$27.49$&$28.19$&$16.20$&$16.00$&$59.19$&$59.68$&$49.88$&$49.57$&$37.85$&$40.62$&$6.16$\\[0.25em]\hdashline\\[-0.7em]

\multicolumn{16}{c}{\textit{Performance using different methods to quantify the similarity between token representations. }} \\[0.5em]\hdashline\\[-0.4em]

$\text{Cosine Sim.}$
&$24.92$&$25.17$&$51.25$&$52.97$&$25.73$&$25.81$&$17.40$&$18.20$&$51.96$&$53.26$&$49.72$&$49.17$&$36.83$&$37.43$&$10.73$\\

$\text{CSLS Sim.}$
&$26.47$&$27.06$&$56.39$&$49.91$&$26.05$&$26.31$&$16.40$&$16.60$&$53.48$&$53.75$&$49.88$&$48.62$&$38.11$&$37.04$&$7.49$\\

$\text{MUSE}$
&$39.77$&$41.88$&$59.30$&$53.91$&$30.51$&$30.74$&$18.20$&$16.00$&$58.27$&$58.71$&$51.14$&$52.96$&$42.87$&$42.37$&$4.51$\\

$\text{VecMap}$
&$\textbf{48.11}$&$\textbf{50.00}$&$59.24$&$55.08$&$\textbf{32.24}$&$\textbf{32.17}$&$18.80$&$\textbf{18.60}$&$\textbf{63.00}$&$\textbf{62.79}$&$\textbf{53.83}$&$\textbf{53.67}$&$\textbf{45.87}$&$\textbf{45.38}$&$\textbf{4.15}$\\

\bottomrule[1.2pt]
\end{tabular}
\end{table*}

\paragraph{Comparison between Different Pooling Methods} Given the hidden states of the last layer of large language models, there are three common methods to obtain the final token representation: max pooling, average pooling, and the hidden state of the last token. 
As shown in Table \ref{tab:ablation_res}, the last token representation method is better than the other methods to obtain the token representation for alignment. 
It may be due to the last token prediction pre-training method of large language models contributing to better last token representations. 
In order to investigate the impact of model parameter amount, we use the hidden states of Pythia${}_{\text{6.9B}}$ in the initialization of Pythia${}_{\text{1B}}$, achieving an inferior performance than its own hidden states. 
Therefore, we recommend adopting the hidden states themselves in the token alignment stage for better initialization performance. 

\paragraph{Bilingual Lexicon Induction Methods} Table \ref{tab:ablation_res} reports the performance of initialized models with different bilingual lexicon induction methods. 
As expected, CSLS similarity performs better than naive cosine similarity by mitigating the hubness problem in the high-dimensional space. 
Sophisticated bilingual lexicon induction methods like MUSE~\cite{lample2018word} further advance the performance of initialized models. 
MUSE relies on an adversarial training paradigm, where a generator and a discriminator are trained iteratively to learn a matrix projecting vectors into the shared space. 
In line with the previous study~\cite{glavas-etal-2019-properly} in the BLI task, VecMap is better than MUSE in the token alignment task, achieving the best results on the language modeling task and downstream benchmarks. 
It recovers 92.25\% performance of base model Pythia${}_{\text{1B}}$ without any tuning, and is set to our default token alignment method. 

\begin{figure}[th]
    \centering
    \subfigure[Pythia${}_{\text{1B}}$]{\includegraphics [scale=0.35]{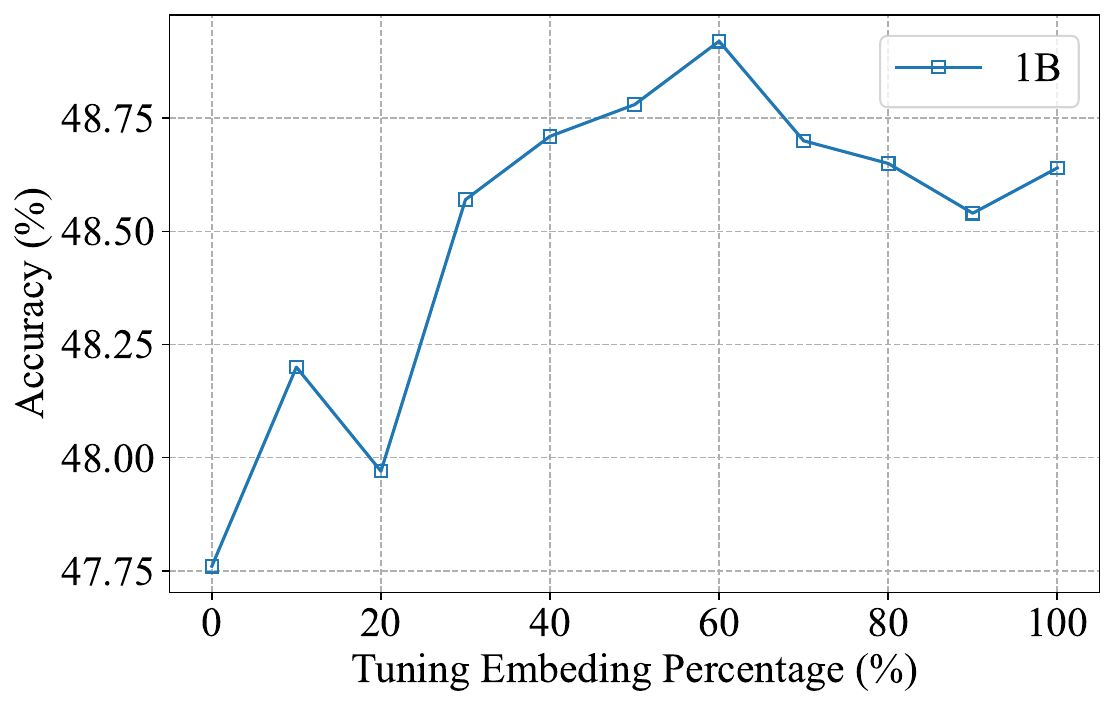}\label{fig:2stage-p1b}}
    \subfigure[Pythia${}_{\text{6.9B}}$]{\includegraphics [scale=0.35]{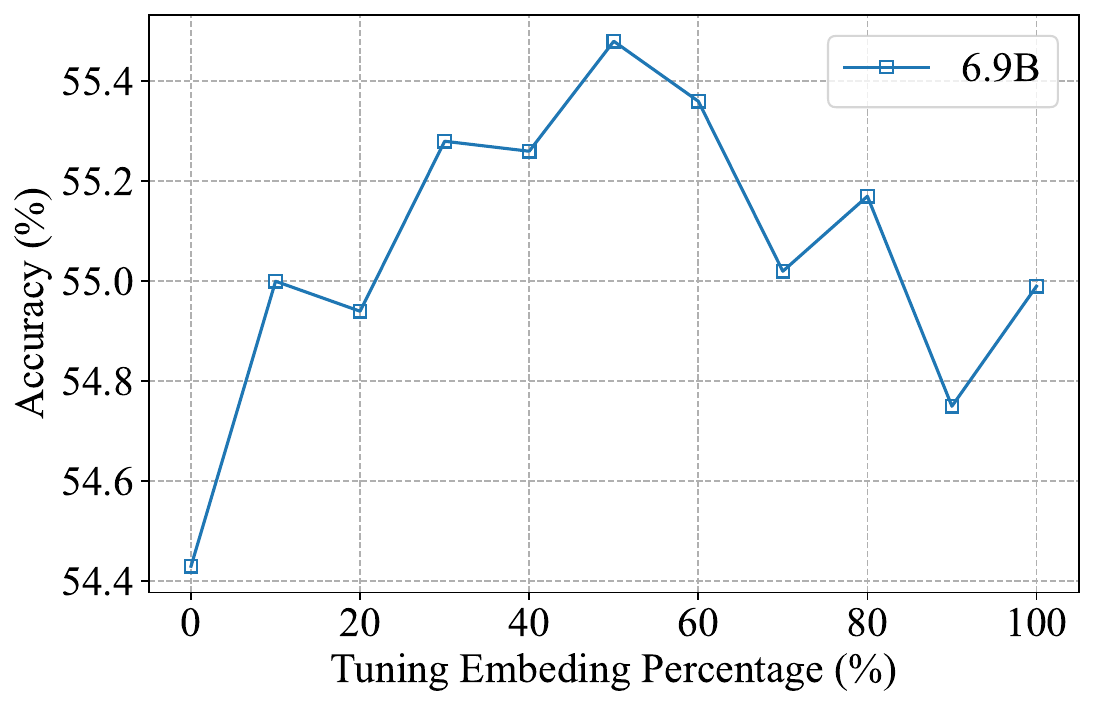}\label{fig:2stage-p6.9b}}
    \caption{\label{fig:2stage_percentage}Influence of the percentage of embedding only tuning stage in the vocabulary adaptation process.}
\end{figure}

\begin{figure*}[th]
    \centering
    \subfigure[Learning rate $\text{4e}^{\text{-5}}$]{\includegraphics [scale=0.3]{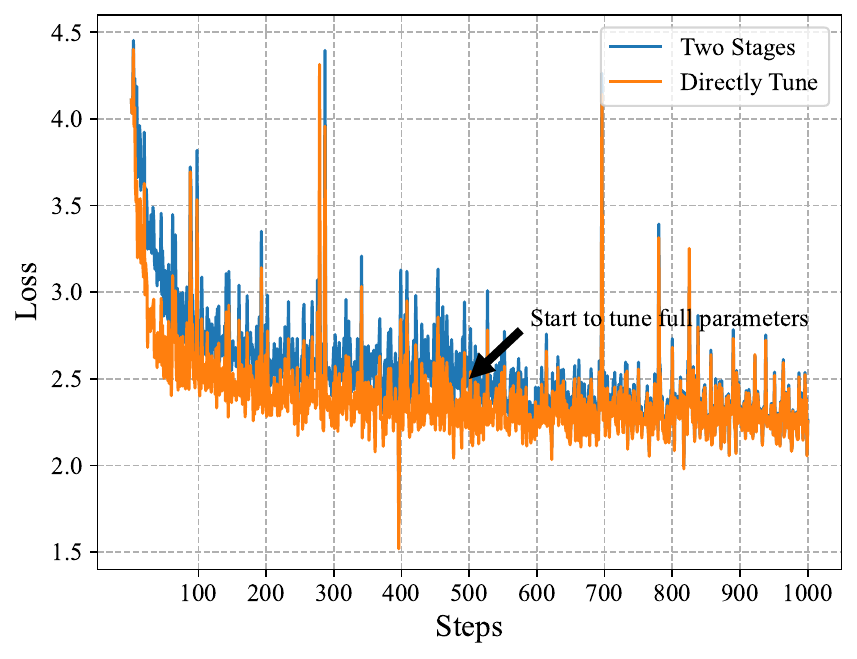}\label{fig:lr4e-5}}
    \subfigure[Learning rate $\text{1.6e}^{\text{-4}}$]{\includegraphics [scale=0.3]{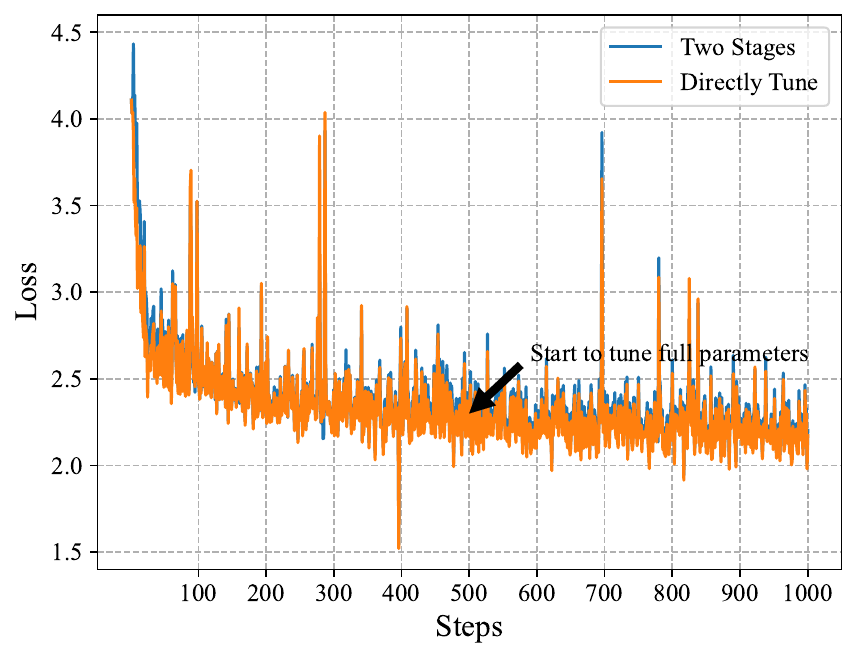}\label{fig:lr1.6e-4}}
    \subfigure[Learning rate $\text{6.4e}^{\text{-4}}$]{\includegraphics [scale=0.3]{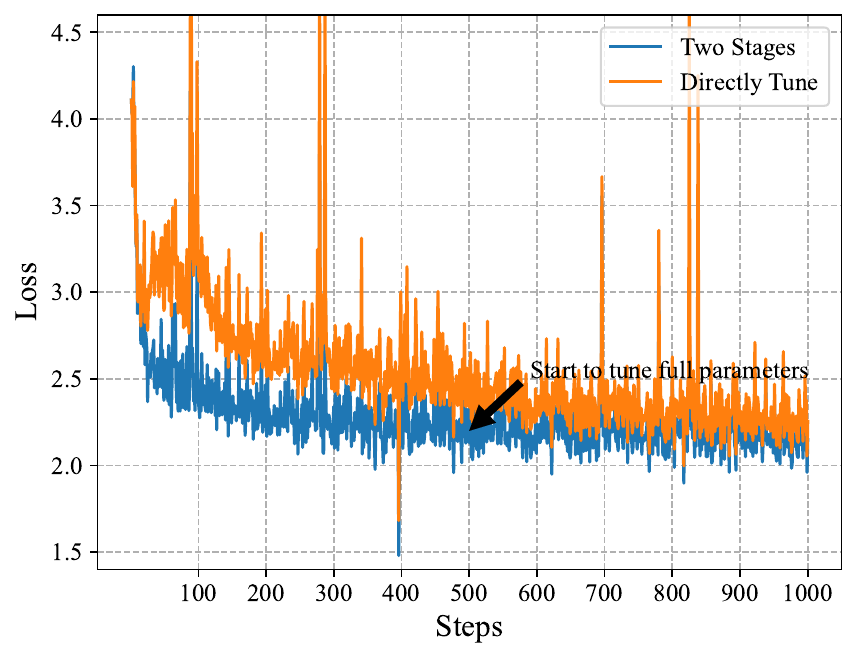}\label{fig:lr6.4e-4}}
    \caption{\label{fig:2stage_loss}The loss curve of Pythia${}_{\text{1B}}$ under two-stage tuning or direct full parameters tuning. }
\end{figure*}

\paragraph{Progressively Vocabulary Adaptation} We modify the percentage of embedding-only fine-tuning to investigate its impact on the final performance after two-stage vocabulary adaptation. 
As shown in Figure \ref{fig:2stage_percentage}, it can be found that the model performs better when the percentage of fine-tuning is set to 50\% or 60\%. 
The fully parameter fine-tuning (0\%) is inferior to the fully vocabulary-related parameters fine-tuning (100\%) for 1B and 6.9B models. 
The reason may be that the newly initialized parameters, which are embeddings and language modeling head parameters, affect the well-trained internal parameters if fine-tuning them at first. 
Our two-stage tuning method can also train with a higher learning rate and less loss spike, like Figure \ref{fig:2stage_loss}, achieving faster convergence and better vocabulary adaptation results. 

\begin{table*}[htp]

\renewcommand\arraystretch{1.3}

\centering
\scriptsize

\setlength{\tabcolsep}{2.2mm}
\caption{\label{tab:multilingual_case} The case study of new Gemma tokens from six languages in the target vocabulary with top-3 source tokens aligned. 
The language families of English, Chinese, and Korean are Indo-European, Sino-Tibetan, and Koreanic, respectively. 
}
 \begin{tabular}{lcccccc}
 
 \toprule[1.2pt]
  
  \multicolumn{1}{c}{ } & \multicolumn{3}{c}{\textbf{English}} & \multicolumn{3}{c}{\textbf{Chinese}} \\

  \cmidrule(r){2-4} \cmidrule(r){5-7}  \noalign{\smallskip}

 \multicolumn{1}{c}{\textbf{Top-3}} & \textbf{\_legend} & \textbf{\_mingled} & \textbf{\_nclui} & \textbf{\begin{CJK*}{UTF8}{gbsn}占比\end{CJK*} (proportion)} & \textbf{\begin{CJK*}{UTF8}{gbsn}分辨率\end{CJK*} (resolution)} & \textbf{\begin{CJK*}{UTF8}{gbsn}频率\end{CJK*} (frequency)} \\

 

\midrule[0.8pt]




   

$1$&Ġlegend&ograft&Ġcores&\%,&1080&Hz\\

$2$&legend&Ġdealings&Ġvaried&\%-&768&nm\\

$3$&Ġneuron&Ġcomposing&oque&\%)&MB&WM\\

\midrule[0.8pt]

\multicolumn{1}{c}{ } & \multicolumn{3}{c}{\textbf{Korean}} & \multicolumn{3}{c}{\textbf{Thai}} \\

  \cmidrule(r){2-4} \cmidrule(r){5-7}  \noalign{\smallskip}

 \multicolumn{1}{c}{\textbf{Top-3}} & \textbf{\begin{CJK}{UTF8}{mj}땐\end{CJK} (time)} & \textbf{\begin{CJK}{UTF8}{mj}각주\end{CJK} (footnote)} & \textbf{\begin{CJK}{UTF8}{mj}굵\end{CJK} (bold)} &  \includegraphics [scale=0.1]{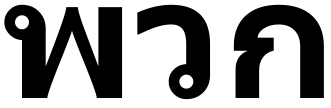} \textbf{(they)} & \includegraphics [scale=0.1]{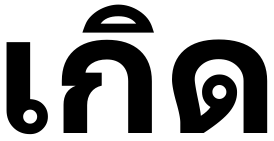} \textbf{(happen)} & \includegraphics [scale=0.11]{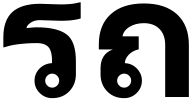} \textbf{(car)} \\


\midrule[0.8pt]

$1$&Asked&unreadable&ĠNBA&they&Ġexpired&Ġretrograde\\

$2$&Ġrecited&hether&ĠMVP&à¹Ģ&ãģ¹&à¹Ģ\\

$3$&Ġstops&.[\^&Ġbooster&ãĤi&Ġassignment&370\\

\midrule[0.8pt]

\multicolumn{1}{c}{ } & \multicolumn{3}{c}{\textbf{Tamil}} & \multicolumn{3}{c}{\textbf{Urdu}} \\

\cmidrule(r){2-4} \cmidrule(r){5-7}  \noalign{\smallskip}

\multicolumn{1}{c}{\textbf{Top-3}} & \includegraphics [scale=0.09]{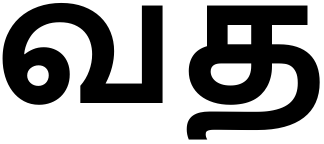} \textbf{(vat)} & \includegraphics [scale=0.09]{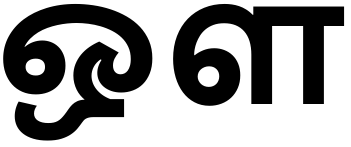} \textbf{(oh)} & \includegraphics [scale=0.07]{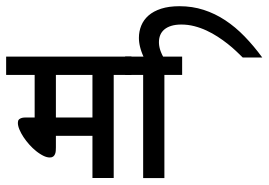} \textbf{(min)} & \textbf{\_}\includegraphics [scale=0.08]{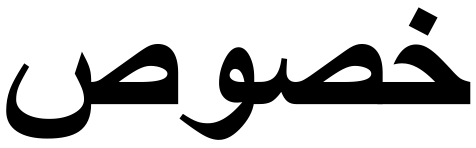} \textbf{(special)} & \textbf{\_}\includegraphics [scale=0.09]{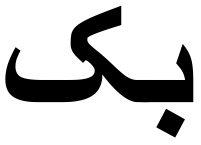} \textbf{(book)} & \includegraphics [scale=0.09]{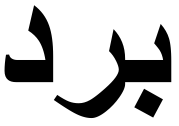} \textbf{(win)} \\

\midrule[0.8pt]

$1$&Ġpaw&lesh&Ġ457&Ġfeatured&ductor&alted\\

$2$&Ġselected&Ġleave&Ġ399&Ġviews&Ġfollowers&Ġdart\\

$3$&Ġstripes&ĠComments&Ġ314&Ġhelm&Ġ1400&Ġ\$\{\{\\

\bottomrule[1.2pt]
\end{tabular}


\end{table*}

\begin{figure*}[th]
    \centering
    \subfigure[1B Embedding]{\includegraphics [scale=0.3]{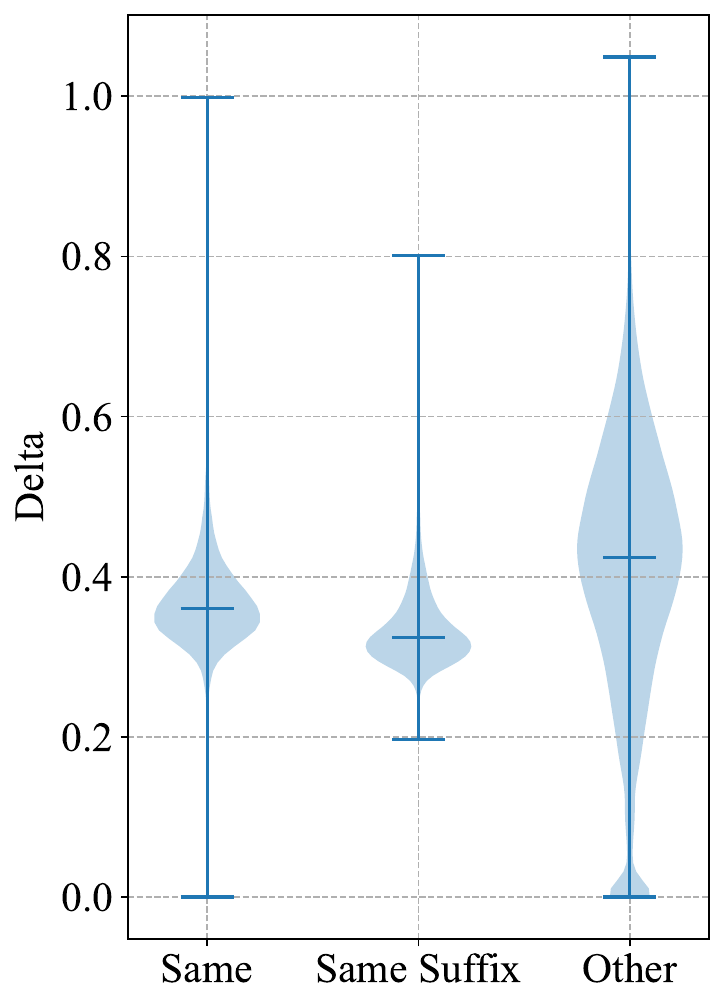}\label{fig:delta_stat_p1b_emb}}
    \subfigure[2.8B Embedding]{\includegraphics [scale=0.3]{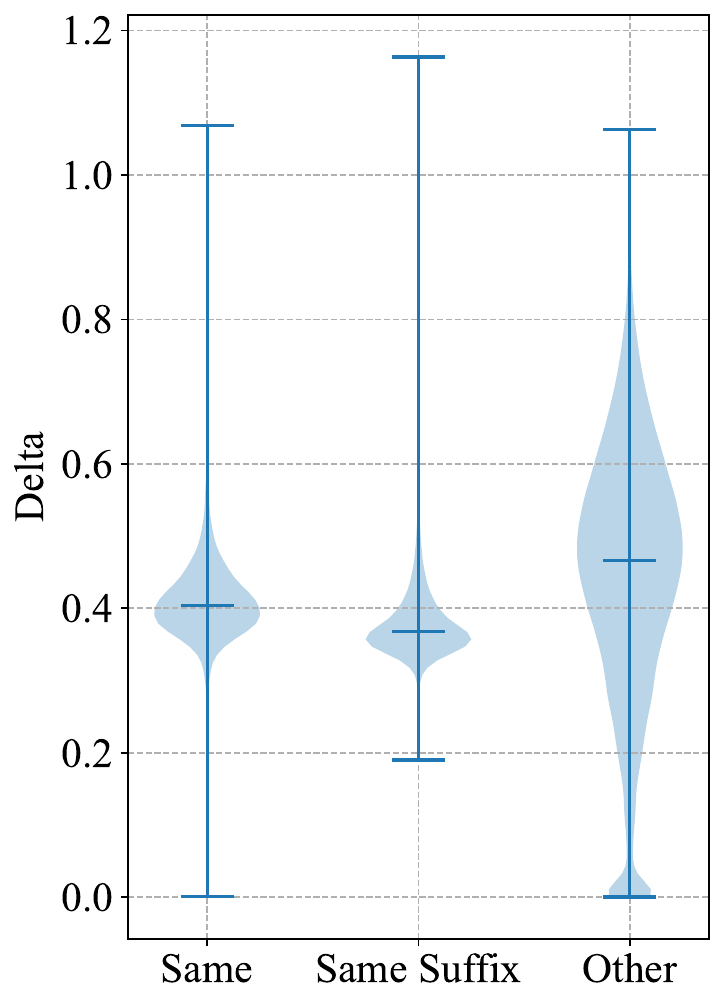}\label{fig:delta_stat_p2b8_emb}}
    \subfigure[6.9B Embedding]{\includegraphics [scale=0.3]{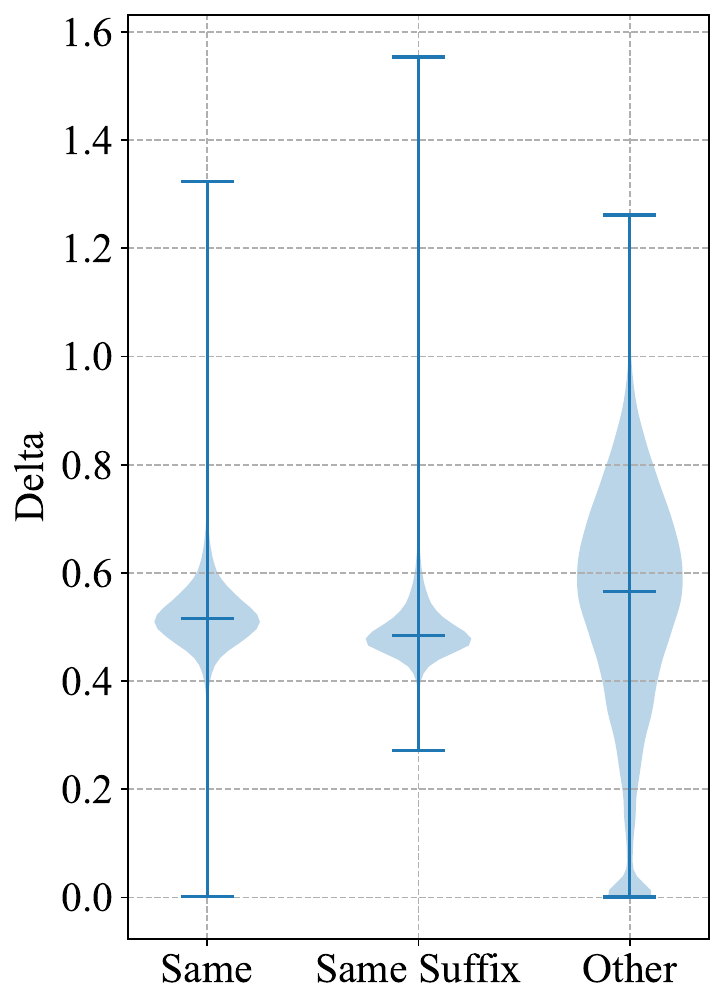}\label{fig:delta_stat_p6b9_emb}}
    \subfigure[1B LM\_Head]{\includegraphics [scale=0.3]{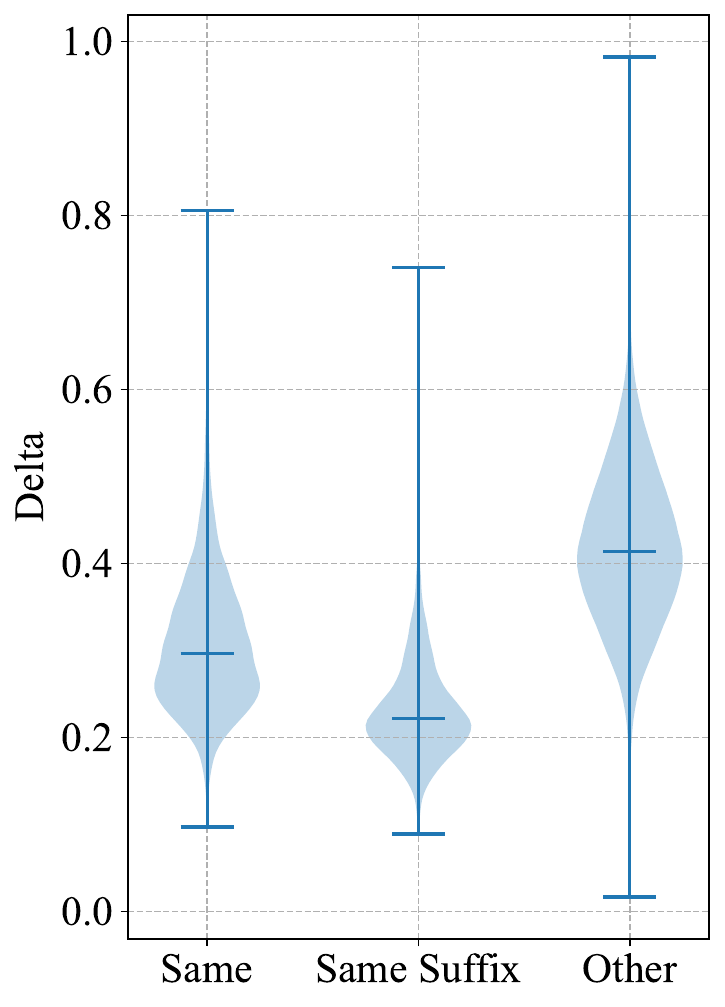}\label{fig:delta_stat_p1b_lm}}
    \subfigure[2.8B LM\_Head]{\includegraphics [scale=0.3]{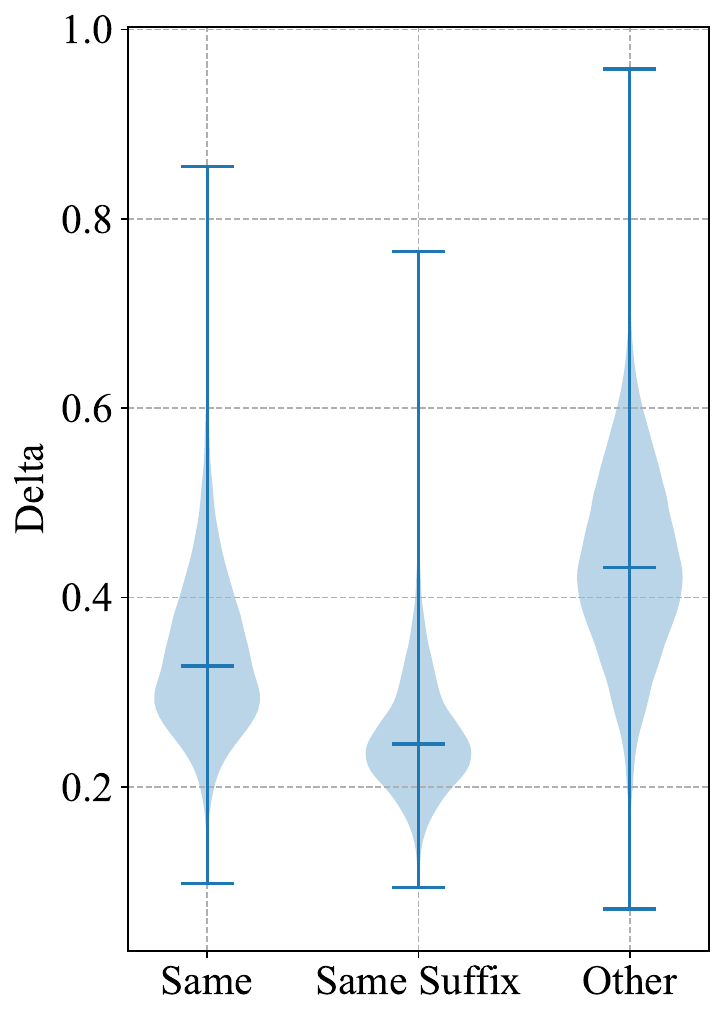}\label{fig:delta_stat_p2b8_lm}}
    \subfigure[6.9B LM\_Head]{\includegraphics [scale=0.3]{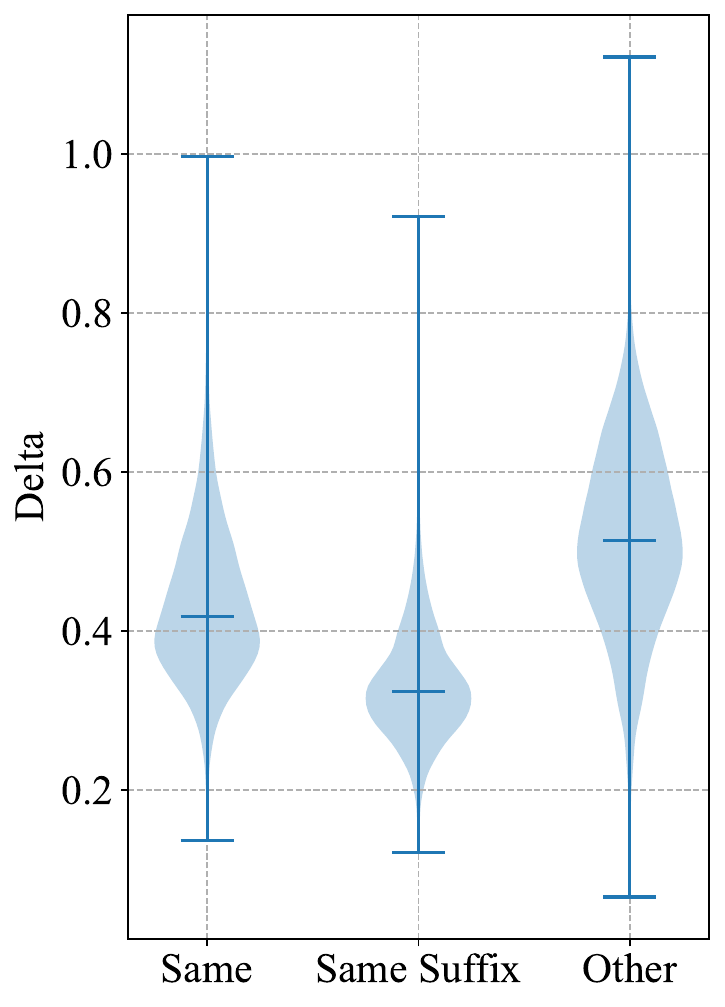}\label{fig:delta_stat_p6b9_lm}}
    \caption{\label{fig:delta_stat_pythia} The statistic of parameter deviation in Pythia models after vocabulary adaptation.}
\end{figure*}

\subsection{Case Study}
To provide a more intuitive demonstration of the effectiveness of TokAlign++ in cross-lingual vocabulary adaptation, we conducted a case study on the alignment of multilingual tokens in the Gemma vocabulary.
Table \ref{tab:multilingual_case} presents the new Gemma tokens from six languages in the target vocabulary along with their top-3 aligned source tokens. These tokens cover English, Chinese, Korean, Thai, Tamil, and Urdu, thereby enabling an evaluation of the method’s alignment capability across high-, medium-, and low-resource languages.

For high-resource languages, TokAlign++ often aligns new tokens in the target vocabulary with source tokens that are semantically or morphologically similar. For example, the English token ``\_legend'' is almost perfectly matched with ``Ġlegend'' in both semantics and morphology, while ``\_mingled'' shows semantic proximity to ``ograft''. In Chinese, tokens such as ``\begin{CJK*}{UTF8}{gbsn}占比\end{CJK*} (proportion)'', ``\begin{CJK*}{UTF8}{gbsn}分辨率\end{CJK*} (resolution)'', and ``\begin{CJK*}{UTF8}{gbsn}频率\end{CJK*} (frequency)'' are aligned with tokens related to percentage, resolution units, and frequency units, respectively. Interestingly, although these tokens are not paired with strictly parallel semantic counterparts, they nevertheless exhibit meaningful associations. These cases illustrate that TokAlign++ can effectively capture semantic consistency in high-resource languages.

For medium-resource languages, TokAlign++ is still able to identify relatively appropriate source tokens for initialization. For instance, the Thai token ``\includegraphics [scale=0.1]{imgs/case_lang/Thai1.pdf} (they)'' is precisely aligned with the English token ``they''. However, not all results exhibit strict semantic parallelism; in some cases, the alignments reflect associations with related attributes or behaviors. For example, the Korean token ``\begin{CJK}{UTF8}{mj}굵\end{CJK} (bold)'' is aligned with ``ĠNBA'', and the Thai token ``\includegraphics [scale=0.11]{imgs/case_lang/Thai3.pdf} (car)'' is aligned with ``Ġretrograde''. Although these results are not perfectly parallel in semantics, they remain within a reasonable semantic or contextual neighborhood, indicating that TokAlign++ maintains robustness in medium-resource settings.

For low-resource languages, the method can still produce several reasonable alignments. For example, the Urdu token ``\textbf{\_}\includegraphics [scale=0.08]{imgs/case_lang/Urdu1.pdf} (special)'' is matched with ``Ġfeatured''. Nevertheless, for some tokens in low-resource languages, the alignment is still not sufficiently intuitive in terms of semantics, and future work could further improve such alignments.

In summary, the results in Table \ref{tab:multilingual_case} demonstrate that TokAlign++ is capable of achieving semantically reasonable alignments across languages of varying resource levels. This provides a reliable foundation for parameter initialization with new vocabularies, thereby ensuring the adaptability and generalization ability of the model in multilingual tasks.

We further examined the parameter deviations of the embedding and LM\_Head after vocabulary adaptation across three categories of tokens: (1) tokens that remain exactly identical before and after alignment, (2) tokens sharing identical suffixes like ``\_legend'' and ``Ġlegend'', and (3) all other tokens. The results, as illustrated in Figure \ref{fig:delta_stat_pythia}, indicate that for models of different scales (1B, 2.8B, and 6.9B), the parameter deviations of Embedding and LM\_Head corresponding to identical tokens and suffix-sharing tokens are generally smaller than those of other tokens. This finding suggests that tokens with higher alignment consistency require less adjustment during the subsequent vocabulary adaptation process. Consequently, high-quality vocabulary alignment initialization plays a crucial role in vocabulary adaptation, underscoring the significance of our work.

\section{Limitations and Future Work}
\label{sec:limitations}
The main limitation of TokAlign++ lies in the 1k fine-tuning steps to adapt the new initialized parameters, which brings additional computational cost. 
It does not achieve the comparable performance as the original model after initialization.
Therefore, the method should be viewed as a high-efficiency initialization strategy rather than a training-free tokenizer transfer method.

Another limitation of TokAlign++ is the performance degradation when the target vocabulary size is substantially smaller than the source (e.g., adapting Pythia-50.3k to LLaMA-32k). 
The reduction in the number of embedding parameters results in a loss of pre-trained information that cannot be fully compensated for by the parameter rearrangement and progressive adaptation process. 

We acknowledge that the efficacy of token alignment lexicon is coupled with the quality and diversity of the corpora used for representation learning. 
Tokens appearing in the ``long tail'' of the distribution often suffer from poor representation due to insufficient co-occurrence data. 
Consequently, the alignment for rare or domain-specific tokens may be suboptimal, potentially impacting the performance of model in specialized tasks unless a balanced and comprehensive corpus is provided.

Future work could explore more challenging scenarios like zero-shot vocabulary adaptation. 
Moreover, TokAlign++ focuses on the vocabulary adaptation problem of large language models, which can be easily extended to other domains, like multi-modal, in the future. 

\section{Conclusion}
\label{sec:conclusion}
In this work, we have proposed TokAlign++, an efficient vocabulary adaptation method for large language models. 
It surpasses the previous method, TokAlign, by improving the alignment between the source and target vocabularies, and provides another lower-cost method to obtain token representations from the vanilla model. 
Experimental results on open-source and open-weight large language models across ten benchmarks demonstrate the superiority of TokAlign++ over the other baseline methods. 

\appendix

\appendixsection{Tokenizer Compression Rate}

Table \ref{tab:tok} reports detailed compression rates of tokenizers across different domains and languages. 
We randomly sample 10 subsets or languages from vanilla datasets \cite{azerbayev2024llemma, kocetkov2023stack} to estimate the compression rate. 
Following \citet{lai-etal-2023-okapi}, the division of languages between ``High'', ``Medium'' and ``Low'' is determined by the available amount resource on CommonCrawl. 
It can be found that vanilla tokenizers often perform inefficiently on specific domains or low-resource languages like Armenian and Icelandic, leading to longer token sequences that slow down training and inference. 

\begin{table*}[htp]

\renewcommand\arraystretch{1.05}

\centering
\small

\setlength{\tabcolsep}{1.2mm}
\caption{\label{tab:tok} The compression rates (bytes/token) of different tokenizers.
}
 \begin{tabu}{c|c|ccccc}
 
 \toprule[1.2pt]
  
  \multicolumn{2}{c}{ } & \multicolumn{5}{c}{\textbf{Tokenizer}} \\

  \cmidrule(r){3-7} \noalign{\smallskip}

 \multicolumn{1}{c}{\textbf{Domain}} & \multicolumn{1}{c}{\textbf{Subset / Language}} & \multicolumn{1}{c}{\textbf{Gemma}} & \textbf{LLaMA3} & \textbf{LLaMA2} & \textbf{Qwen2} & \textbf{Pythia} \\

\midrule[0.8pt]
\multirow{5}{*}{\makecell{\textbf{Math} \\ \citep{azerbayev2024llemma}}} &
$ArXiv$&$2.8561$&$2.7765$&$2.7040$&$2.7445$&$2.8489$ \\
&$Textbooks$&$4.0883$&$4.3270$&$3.6500$&$4.2899$&$3.9464$ \\
&$Wikipedia$&$3.1753$&$3.2049$&$2.8792$&$3.0312$&$3.2898$ \\
&$ProofWiki$&$2.7538$&$2.8115$&$2.5996$&$2.7900$&$2.7363$ \\
&$StackExchange$&$3.2062$&$3.2814$&$3.0094$&$3.2107$&$3.2222$ \\
&$WebPages$&$3.9885$&$4.0655$&$3.5070$&$3.8720$&$4.1136$ \\[0.25em]\hdashline\\[-0.5em]

\multirow{10}{*}{\makecell{\textbf{Code} \\ \citep{kocetkov2023stack}}} &
$Python$&$3.3401$&$4.1331$&$3.0072$&$4.0339$&$3.2328 $ \\
&$Java$&$3.7175$&$4.4900$&$3.2193$&$4.4141$&$3.4914 $ \\
&$Go$&$2.9274$&$3.4797$&$2.5189$&$3.3870$&$2.8542 $ \\
&$VHDL$&$2.1038$&$2.4814$&$1.8724$&$2.2961$&$2.1395 $ \\
&$ActionScript$&$3.3470$&$3.9717$&$2.7852$&$3.9180$&$3.2949 $ \\
&$Scheme$&$2.7178$&$3.3045$&$2.4586$&$2.9713$&$2.9326 $ \\
&$Haml$&$3.2423$&$3.8429$&$2.9588$&$3.8002$&$3.1016 $ \\
&$Xbase$&$2.8739$&$3.4325$&$2.3300$&$3.3475$&$2.7837 $ \\
&$Mako$&$3.4387$&$4.0746$&$3.1238$&$4.0311$&$3.2844 $ \\
&$EmberScript$&$1.4104$&$1.9017$&$1.3819$&$1.4082$&$2.1540 $ \\[0.25em]\hdashline\\[-0.5em]

\multirow{10}{*}{\makecell{\textbf{High-Langs} \\ \citep{nguyen2023culturax}}} &
$English$&$4.4971 $&$4.6042 $&$3.8647 $&$4.4875 $&$4.4505 $ \\
&$Russian$&$6.7529 $&$5.8131 $&$4.9275 $&$5.3559 $&$3.5802 $ \\
&$Spanish$&$4.6068 $&$3.8416 $&$3.4517 $&$3.8330 $&$3.3655 $ \\
&$German$&$4.4605 $&$3.6314 $&$3.4417 $&$3.6041 $&$3.1096 $ \\
&$French$&$4.2258 $&$3.7378 $&$3.4445 $&$3.7243 $&$3.3565 $ \\
&$Chinese$&$3.7378 $&$3.2373 $&$1.8434 $&$3.9859 $&$1.9896 $ \\
&$Italian$&$4.2211 $&$3.4952 $&$3.3320 $&$3.4573 $&$3.1928 $ \\
&$Portuguese$&$4.2731 $&$3.6030 $&$3.2031 $&$3.5850 $&$3.2022$ \\ 
&$Polish$&$3.5583 $&$2.8548 $&$2.6639 $&$2.9464 $&$2.4333 $ \\
&$Japanese$&$5.7640 $&$4.2796 $&$2.4701 $&$4.7059 $&$2.9326$ \\[0.25em]\hdashline\\[-0.5em]

\multirow{10}{*}{\makecell{\textbf{Medium-Langs}\\ \citep{nguyen2023culturax}}} &
$Czech$&$3.3402 $&$3.2875 $&$2.5978 $&$2.4490 $&$2.3884 $ \\
&$Vietnamese$&$4.5376 $&$4.2766 $&$1.9699 $&$4.2877 $&$2.0382 $ \\
&$Persian$&$5.6465 $&$5.3015 $&$1.7938 $&$3.1923 $&$2.3707 $ \\
&$Hungarian$&$3.2337 $&$2.6008 $&$2.6311 $&$2.5500 $&$2.3878 $ \\
&$Greek$&$4.4691 $&$4.5671 $&$1.8544 $&$2.1225 $&$3.0283 $ \\
&$Romanian$&$3.5558 $&$3.0566 $&$2.8355 $&$3.0083 $&$2.8981 $ \\
&$Swedish$&$3.7087 $&$3.1398 $&$2.9214 $&$3.0977 $&$2.9620 $ \\
&$Ukrainian$&$5.5141 $&$5.5985 $&$4.5904 $&$3.6179 $&$3.0702 $ \\
&$Finnish$&$3.2659 $&$2.6748 $&$2.4176 $&$2.6473 $&$2.6112 $ \\
&$Korean$&$3.3556 $&$3.6957 $&$1.5977 $&$3.3330 $&$1.5667 $ \\[0.25em]\hdashline\\[-0.5em]

\multirow{10}{*}{\makecell{\textbf{Low-Langs}\\ \citep{nguyen2023culturax}}} &
$Hebrew$&$4.0487 $&$1.8592 $&$1.7875 $&$4.3773 $&$2.0380 $ \\
&$Serbian$&$4.8596 $&$3.9234 $&$4.2642 $&$3.6267 $&$2.9896 $ \\
&$Tamil$&$5.6161 $&$2.0279 $&$2.2615 $&$2.4759 $&$1.9765 $ \\
&$Albanian$&$2.8919 $&$2.6536 $&$2.2945 $&$2.6037 $&$2.3631 $ \\
&$Azerbaijani$&$2.8585 $&$2.4857 $&$2.0407 $&$2.3797 $&$2.1534 $ \\
&$Kazakh$&$3.8172 $&$2.9176 $&$3.0869 $&$2.9263 $&$2.3236 $ \\
&$Urdu$&$4.4364 $&$2.8462 $&$1.7260 $&$2.7174 $&$1.9458 $ \\
&$Georgian$&$3.8237 $&$1.4828 $&$2.5595 $&$2.6951 $&$2.2077 $ \\
&$Armenian$&$3.2133 $&$1.1658 $&$1.7000 $&$1.8531 $&$1.3922 $ \\
&$Icelandic$&$2.7964 $&$2.4860 $&$2.3050 $&$2.4330 $&$2.3185 $ \\

\bottomrule[1.2pt]
\end{tabu}

\end{table*}

\appendixsection{Language Codes}

We provide details of 15 languages in Table \ref{tab:lang_codes}. 
These languages come from 9 language families and covers three types of amount resource on CommonCrawl. 

\begin{table}[htp]

\setlength{\tabcolsep}{4mm}
\centering
\renewcommand\arraystretch{1.2}
\caption{\label{tab:lang_codes} Details of language codes in this work. 
}
\begin{center}
    \begin{tabular}{ccc}
        \toprule[1.2pt] 
        \multicolumn{1}{c}{\textbf{ISO 639-1}} & \multicolumn{1}{c}{\textbf{Language}}    &\multicolumn{1}{c}{\textbf{Family}} \\
                \midrule[0.8pt]
                ar      &Arabic      &Afro-Asiatic\\
                bn      &Bengali      &Indo-European\\
                de      &German      &Indo-European\\
                en      &English      &Indo-European\\
                ja      &Japanese      &Japonic\\
                ko      &Korean      &Koreanic\\
                ml      &Malayalam   &Dravidian\\
                mn      &Mongolian   &Mongolic\\
                ta      &Tamil      &Dravidian\\
                te      &Telugu      &Dravidian\\
                th      &Thai      &Kra-Dai\\
                uk      &Ukrainian  &Indo-European\\
                ur      &Urdu      &Indo-European\\
                vi      &Vietnamese      &Austroasiatic\\
                zh      &Chinese      &Sino-Tibetan\\
        \bottomrule[1.2pt]
    \end{tabular}
\end{center}

\end{table}

\begin{acknowledgments}
We would like to thank the anonymous reviewers and editors for their efforts and valuable comments.
This work was supported by the National Key R\&D Program of China (No. 2022ZD0160602) and the
Strategic Priority Research Program of Chinese Academy of Sciences (No. XDA04080400).
\end{acknowledgments}


\bibliographystyle{compling}
\bibliography{COLI_template}

\end{document}